%% file: main.tex
\documentclass{article}[11pt]

\usepackage{microtype}
\usepackage{graphicx}
\usepackage{subfigure}
\usepackage{booktabs} 
\usepackage{makecell}
\usepackage{subfigure}
\usepackage{paralist}
\usepackage[usenames,dvipsnames]{xcolor}

\usepackage[colorlinks=true,citecolor=blue]{hyperref}       

\usepackage[left=1.25in, top=1in, bottom=1in, right=1.25in]{geometry}
\title{Does label smoothing mitigate label noise?}
\author{Michal Lukasik, Srinadh Bhojanapalli, Aditya Krishna Menon and Sanjiv Kumar \\
Google Research, New York \\
\texttt{\{mlukasik,bsrinadh,adityakmenon,sanjivk\}@google.com}}

\input{macros.tex}
\usepackage{cleveref}

\begin{document}

\maketitle

%
\input{body}

\bibliography{references}
\bibliographystyle{plainnat}

\clearpage
\appendix
\onecolumn

\input{appendix}

\end{document}

%% file: macros.tex
\usepackage[utf8]{inputenc} 
\usepackage[T1]{fontenc}    
\usepackage{url}            
\usepackage{booktabs}       
\usepackage{amsfonts}       
\usepackage{nicefrac}       
\usepackage{microtype}      

\usepackage{natbib}

\usepackage{float}
\usepackage{graphicx}
\usepackage[export]{adjustbox}
\usepackage[inline]{enumitem}

\usepackage{tikz}
\usetikzlibrary{positioning}

\usepackage{amsmath,amssymb,amsthm}
\usepackage[mathscr]{eucal}
\usepackage{stmaryrd}
\usepackage{accents}
\usepackage{upgreek}

\usepackage[most]{tcolorbox}
\usepackage{setspace}

\usepackage{booktabs,colortbl,multirow}

\usepackage{algorithm,algorithmic}

\graphicspath{{figs//}}

\allowdisplaybreaks

\newcommand\numberthis{\addtocounter{equation}{1}\tag{\theequation}}

\newcommand{\best}[1]{\cellcolor{gray!25}{#1}}
\newcommand{\pStar}{\mathbf{p}^*}
\newcommand{\pSmearStar}{\mathbf{p}^{\mathrm{SM}}}
\newcommand{\tup}{\mathrm{T}}
\newcommand{\NoiseMat}{\mathbf{T}}
\newcommand{\Identity}{\mathbf{I}}
\newcommand{\Ones}{\mathbf{J}}
\newcommand{\SmearMatrix}{\mathbf{M}}
\newcommand{\ellBack}{\ell^{\leftarrow}}
\newcommand{\ellForward}{\ell^{\rightarrow}}
\newcommand{\rhoPrime}{\alpha}

\newcommand{\defEq}{\stackrel{.}{=}}

\newcommand{\indicator}[1]{\llbracket #1 \rrbracket}

\renewcommand{\Pr}{\mathbb{P}}
\newcommand{\E}[2]{\underset{#1}{\mathbb{E}}\left[ #2 \right]}

\newcommand{\X}{x}
\newcommand{\Y}{y}

\newcommand{\XCal}{\mathscr{X}}
\newcommand{\YCal}{\mathscr{Y}}

\newcommand{\Real}{\mathbb{R}}

\newtheorem{lemma}{Lemma}

\newtheorem{theorem}[lemma]{Theorem}

\theoremstyle{definition}

\newtcbtheorem[number within=section]%
{remark_tcb} 
{Remark} 
{%
fonttitle=\bfseries\upshape,fontupper=\normalsize,
colframe=green!10!black,colback=green!2!white,
colbacktitle=green!20!white,coltitle=blue!75!black,
boxsep=1pt,left=2pt,right=2pt,top=1pt,bottom=1pt,
frame hidden,boxrule=1pt,
theorem style=plain} 
{rem} 

\newtcbtheorem[number within=section]%
{assumption_tcb} 
{Assumption} 
{%
fonttitle=\bfseries\upshape,fontupper=\normalsize,
colframe=blue!5!black,colback=blue!2!white,
colbacktitle=blue!10!white,coltitle=blue!75!black,
boxsep=1pt,left=2pt,right=2pt,top=1pt,bottom=1pt,
frame hidden,boxrule=1pt,
theorem style=plain} 
{asm} 

\newcommand{\inp}[2]{\left\langle #1,#2\right\rangle}
\newcommand{\w}{\mathbf{W}}

\newcommand{\cifarT}{CIFAR-10}
\newcommand{\cifarH}{CIFAR-100}
\newcommand{\imagenet}{ImageNet}
\newcommand{\resnetT}{ResNet-32}
\newcommand{\resnetF}{ResNet-56}
\newcommand{\resnetIM}{ResNet-v2-50}

%% file: body.tex

%
\begin{abstract}
\input{abstract}
\end{abstract}

\section{Introduction}
\label{sec:intro}
\input{intro}

%
\section{Background and notation}
\label{sec:background}
\input{background}

%
\section{Label smoothing meets loss correction}
\label{sec:label_smearing}
\input{label_smearing.tex}
\section{Effect of label smoothing on label noise}
\label{sec:noise}
\input{experiments_denoise.tex}
\input{smoothing_regularisation.tex}

%
\section{Distillation under label noise}
\label{sec:distillation}
\input{noisy_distillation.tex}
\section{Conclusion}
\input{conclusion}

%% file: abstract.tex
Label smoothing is commonly used in training deep learning models,
wherein one-hot training labels are mixed with uniform label vectors.
Empirically, smoothing has been shown to improve both predictive performance and model calibration.
In this paper, we study whether label smoothing is also effective as a means of coping with \emph{label noise}.
While label smoothing apparently \emph{amplifies} this problem 
---
being equivalent to injecting symmetric noise to the labels
---
we show how it relates to a general family of 
\emph{loss-correction} techniques from the label noise literature.
Building on this connection,
we show that 
label smoothing is competitive with loss-correction
under label noise.
Further, we show that when distilling models from noisy data, 
label smoothing of the teacher is beneficial;
this is in contrast to recent findings for noise-free problems,
and sheds further light on settings where label smoothing is beneficial.


%% file: intro.tex

Label smoothing is commonly used to improve the performance of deep learning models~\citep{Szegedy:2016,Chorowski:2017,Vaswani:2017,Zoph:2018,Real:2018,Huang:2019,Li:2020}.
Rather than standard training with one-hot training labels,
label smoothing
prescribes using \emph{smoothed} labels
by mixing in a uniform label vector.
This procedure is generally understood as a means of regularisation~\citep{Szegedy:2016,Zhang:2018} that improves generalization and model calibration~\citep{Pereyra:2017,Muller:2019}.

How does label smoothing affect the robustness of deep networks? 
Such robustness is desirable when
learning from data subject to \emph{label noise}~\citep{Angluin:1988}.
Modern deep networks 
can perfectly fit such noisy labels \citep{Zhang:2017}.
Can label smoothing address this problem?
Interestingly, there are two competing intuitions.
On the one hand, 
smoothing 
might \emph{mitigate} the problem,
as 
it prevents 
overconfidence on any one example.
On the other hand,
smoothing might \emph{accentuate} the problem,
as
it is equivalent to injecting uniform noise into all labels~\citep{Xie:2016}.

Which of these intuitions is borne out in practice?
A systematic study of this question is, to our knowledge, lacking.
Indeed,
label smoothing is conspicuously absent in most treatments of the noisy label problem~\citep{Patrini:2016,Han:2018,Charoenphakdee:2019,Thulasidasan:2019,Amid:2019b}.
Intriguingly, however,
a cursory inspection at popular \emph{loss correction} techniques in this literature~\citep{Natarajan:2013,Patrini:2017,vanRooyen:2018}
reveals a strong similarity to label smoothing (see~\S\ref{sec:label_smearing}).
But what is the precise relationship between these methods, and does it imply label smoothing is a viable denoising technique?

In this paper, we address these questions by first
connecting label smoothing to existing label noise techniques.
At first glance, this connection
indicates that smoothing has an \emph{opposite} effect to one such
\emph{loss-correction} technique.
However,
we empirically show that smoothing is competitive with such techniques in denoising,
and that it 
improves performance of
\emph{distillation}~\citep{Hinton:2015} under label noise.
We 
then explain its denoising ability 
by analysing smoothing
as a \emph{regulariser}.
In sum, our contributions are:
\begin{enumerate}[label=(\roman*),itemsep=-2pt,topsep=-2pt]
    \item we present a novel connection of
    label smoothing to 
    loss correction techniques from the label noise literature~\citep{Natarajan:2013,Patrini:2017}.
    
    \item we empirically demonstrate that 
    label smoothing significantly improves performance under label noise,
    which we explain by relating smoothing to \emph{$\ell_2$ regularisation}.
    
    \item we show that when distilling from noisy labels,
    smoothing the teacher \emph{improves} the student;
    this is in marked contrast to 
    recent findings
    in noise-free settings.
\end{enumerate}
Contributions (i) and (ii) establish that 
label smoothing can be beneficial under noise,
and 
also
highlight that
a \emph{regularisation} view
can complement a
\emph{loss} view,
the latter being more popular in the noise literature~\citep{Patrini:2017}.
Contribution (iii) 
continues a line of exploration initiated in~\citet{Muller:2019}
as to the relationship between teacher accuracy and student performance.
While~\citet{Muller:2019} established that label smoothing can \emph{harm} distillation,
we 
show 
an
\emph{opposite} picture
in noisy settings.



%% file: background.tex
We present some background on 
(noisy) multiclass classification,
label smoothing,
and knowledge distillation.

\subsection{Multiclass classification}

In multiclass classification,
we seek to classify instances $\XCal$
into 
one of $L$ labels $\YCal = [ L ] \defEq \{ 1, 2, \ldots, L \}$.
More precisely,
suppose instances and labels are drawn from a distribution $\Pr$.
Let 
$\ell \colon [ L ] \times \Real^L \to \Real_+$ be a \emph{loss} function, 
where 
$\ell( y, \mathbf{f} )$ is the penalty for predicting \emph{scores} $\mathbf{f} \in \Real^L$ given true label $y \in [ L ]$.
We seek a predictor $\mathbf{f} \colon \XCal \to \Real^L$ minimising the \emph{risk} of $\mathbf{f}$, i.e., its expected {loss} under $\Pr$:
$$ R( \mathbf{f} ) \defEq \E{(\X, \Y)}{ \ell( \Y, \mathbf{f}( \X ) ) } = \E{\X}{ \mathbf{p}^*( \X )^{\tup} \ell( \mathbf{f}( \X ) ) }, $$
where 
$\mathbf{p}^*( x ) \defEq \begin{bmatrix} \Pr( y \mid x ) \end{bmatrix}_{y \in [L]}$ is
the class-probability distribution,
and
$\ell( \mathbf{f} ) \defEq \begin{bmatrix} \ell( y, \mathbf{f} ) \end{bmatrix}_{y \in [L]}$.
Canonically,
$\ell$ is the softmax cross-entropy,
$ \ell( y, \mathbf{f} ) \defEq -f_y + \log \sum_{y' \in [L]} e^{f_{y'}} $.

Given a finite training sample $S = \{ ( x_n, y_n ) \}_{n = 1}^N \sim \Pr^N$, one can minimise the \emph{empirical risk}
$$ R( \mathbf{f}; S ) \defEq \frac{1}{N} \sum_{n = 1}^{N} \ell( y_n, \mathbf{f}( x_n ) ). $$
In \emph{label smoothing}~\citep{Szegedy:2016}, 
one mixes the training labels with a uniform mixture over all possible labels:
for $\alpha \in [0, 1]$,
this corresponds to minimising
\begin{equation}
    \begin{aligned}
    \label{eqn:smoothing}
    \bar{R}( \mathbf{f}; S ) &= \frac{1}{N} \sum_{n = 1}^{N} \bar{\mathbf{y}}_n^{\tup} \ell( \mathbf{f}( x_n ) ),
\end{aligned}
\end{equation}
where $(\bar{\mathbf{y}}_n)_i \defEq (1 - \alpha) \cdot \indicator{ i = y } + \frac{\alpha}{L}$.

%
\subsection{Learning under label noise}

The \emph{label noise} problem is the setting where one observes samples from 
some distribution $\bar{\Pr}$
with $\bar{\Pr}( y \mid x ) \neq \Pr( y \mid x )$;
i.e.,
the observed labels are not reflective of the ground truth~\citep{Angluin:1988,Scott:2013}.
Our goal is to nonetheless minimise the risk on the (unobserved) $\Pr$. %
This poses a challenge to 
deep neural networks, 
which can fit completely arbitrary labels~\citep{Zhang:2017}.

A common means of coping with noise is to posit a noise model,
and design robust procedures under this model.
One simple model is 
\emph{class-conditional noise}~\citep{Blum:1998,Scott:2013,Natarajan:2013},
wherein
there is a row-stochastic \emph{noise transition} matrix $\NoiseMat \in [ 0, 1 ]^{L \times L}$
such that for each $( \X, \Y ) \sim \Pr$,
label $\Y$ may be flipped to $\Y'$ with probability $\NoiseMat_{\Y, \Y'}$.
Formally,
if 
$\bar{\mathbf{p}}^*_y( x ) \defEq \bar{\Pr}( {y} \mid x )$
and
$\mathbf{p}^*_y( x ) \defEq \Pr( y \mid x )$ are the noisy and clean class-probabilities respectively,
we have
\begin{equation}
    \label{eqn:noisy-eta-ccn}
    \bar{\mathbf{p}}^*( x ) = \mathbf{T}^{\tup} \mathbf{p}^*( x ).
\end{equation}
The \emph{symmetric noise} model further assumes that
there is a constant flip probability $\rho \in \left[ 0, 1 - \frac{1}{L} \right)$
of changing the label uniformly to one of the other classes~\citep{Long:2010,vanRooyen:2015},
i.e.,
for $\rhoPrime \defEq \frac{L}{L - 1} \cdot \rho$,
\begin{equation}
    \label{eqn:t-ccn}
    \mathbf{T} = \left( 1 - {\rhoPrime} \right) \cdot \Identity + \frac{\rhoPrime}{L} \cdot \Ones
\end{equation}
where $\Identity$ denotes the identity and $\Ones$ the all-ones matrix.

%

While
there are several approaches to coping with noise,
our interest will be in the family of \emph{loss correction} techniques:
assuming one has knowledge (or estimates) of the noise-transition matrix $\NoiseMat$,
such techniques yield 
consistent risk minimisers with respect to $\Pr$.
\citep{Patrini:2017} proposed two such techniques,
termed \emph{backward} and \emph{forward correction},
which respectively involve the losses
\begin{align}
    \label{eqn:backward-correction}
    \ellBack( \mathbf{f} ) &= \NoiseMat^{-1} \ell( \mathbf{f} ) \\
    \label{eqn:forward-correction}
    \ellForward( \mathbf{f} ) &= \ell( \NoiseMat \mathbf{f} ).
\end{align}
Observe that for a given label $y$, 
$\ellBack( y, \mathbf{f} ) = \sum_{y' \in [L]} T^{-1}_{y y'} \cdot \ell( y', \mathbf{f}( \X ) )$ computes a weighted sum of \emph{losses} for {all} labels $y' \in [L]$, 
while 
$\ellForward( y, \mathbf{f} ) = \ell\left( y, \sum_{y' \in [L]} T_{: y'} \cdot f_{y'}( \X ) \right)$
computes a weighted sum of \emph{predictions} for {all} $y' \in [L]$.

Backward correction was inspired by techniques in~\citet{Natarajan:2013,Cid-Sueiro:2014,vanRooyen:2018},
and results in an unbiased estimate of the risk with respect to $\Pr$.
Recent works have studied robust estimation of the $\mathbf{T}$ matrix from noisy data alone~\citep{Patrini:2017,Han:2018,Xia:2019}.
Forward correction was inspired by techniques in~\citet{Reed:2014,Sukhbaatar:2015},
and 
does \emph{not} result in an unbiased risk estimate.
However, it preserves the Bayes-optimal minimiser,
and is empirically effective~\citep{Patrini:2017}.




%
\subsection{Knowledge distillation}

Knowledge distillation~\cite{Bucilua:2006,Hinton:2015} refers to the following recipe:
given a training sample $S \sim \Pr^N$,
one trains a \emph{teacher} model using a loss function suitable for estimating class-probabilities,
e.g.,
the softmax cross-entropy.
This produces a class-probability estimator $\mathbf{p}^{\mathrm{t}} \colon \XCal \to \Delta_L$,
where $\Delta$ denotes the simplex.
One then uses $\{ ( x_n, \mathbf{p}^{\mathrm{t}}( x_n ) ) \}_{n = 1}^N$
to train a \emph{student} model,
e.g., using cross entropy \cite{Hinton:2015} or square loss \cite{sanh2019distilbert} as an objective.
The key advantage of distillation is that 
the resulting student has improved performance
compared to simply training the student on labels in $S$.



%% file: label_smearing.tex
We 
now relate
label smoothing to
loss correction techniques for label noise
via a \emph{label smearing}
framework.

%
\subsection{Label smearing for loss functions}

Suppose we have some base loss $\ell$ of interest, e.g., the softmax cross-entropy.
Recall that we summarise the loss via the vector $\ell( \mathbf{f} ) \defEq \begin{bmatrix} \ell( y, \mathbf{f} ) \end{bmatrix}_{y \in [L]}$.
The loss on an example $(x, y)$ is $\ell( y, \mathbf{f}( x ) ) = \mathbf{e}_{y}^{\tup} \ell( \mathbf{f}( x ) )$
for one-hot vector $\mathbf{e}_{y}$.

Consider now the following generalisation,
which we term \emph{label smearing}:
given a matrix 
$\SmearMatrix \in \mathbb{R}^{L \times L}$,
we compute
$$ \ell^{\mathrm{SM}}( \mathbf{f} ) \defEq \SmearMatrix \, \ell( \mathbf{f} ). $$
On an example $(x, y)$, the \emph{smeared loss} is given by
\begin{align*}
\mathbf{e}_{y}^{\tup} \ell^{\mathrm{SM}}( \mathbf{f}( x ) )
&= M_{y y} \cdot \ell( y, \mathbf{f}( x ) ) + \sum_{y' \neq y} M_{y y'} \cdot \ell( y', \mathbf{f}( x ) ).
\end{align*}
Compared to the standard loss, we now potentially involve \emph{all} possible labels,
scaled appropriately by the matrix $\SmearMatrix$.

%
\subsection{Special cases of label smearing}

The label smearing framework captures many interesting approaches as special cases (see Table~\ref{tbl:comparison}):
\begin{itemize}[itemsep=0pt,topsep=0pt,leftmargin=12pt]
    \item \emph{Standard training}.
    Suppose that $\SmearMatrix = \Identity$,
    for identity matrix
    $\Identity$.
    This trivially corresponds to standard training.

    \item \emph{Label smoothing}. 
    Suppose that $\SmearMatrix = (1 - \alpha) \cdot \Identity + \frac{\alpha}{L} \cdot \Ones$, where
    $\Ones$ is the all-ones matrix,
    and $\alpha \in [0, 1]$ is a tuning parameter.
This corresponds to mixing the true label with a uniform distribution over all the classes,
which is precisely {label smoothing} per~\eqref{eqn:smoothing}.



    \item \emph{Backward correction}.
    Suppose that $\SmearMatrix = \mathbf{T}^{-1}$,
    where $\mathbf{T}$ is a class-conditional noise transition matrix.
    This corresponds to the backward correction procedure of~\citet{Patrini:2017}.
    Here, the entries of $\SmearMatrix$ may be \emph{negative};
    indeed, for symmetric noise,
    $\SmearMatrix = \frac{1}{1 - \rhoPrime} \cdot \left( \Identity - \frac{\rhoPrime}{L} \cdot \Ones \right)$
    where $\rhoPrime \defEq \frac{L}{L - 1} \cdot \rho$.
    While this poses optimisation problems, 
    recent works have studied means of correcting this~\citep{Kiryo:2017,Han:2018b}.
\end{itemize}

The above techniques have been developed with different motivations.
By casting them in a common framework,
we can elucidate some of their shared properties.

\begin{table}[!t]
    \centering
    \renewcommand{\arraystretch}{1.25}
    \begin{tabular}{@{}ll@{}}
        \toprule
        \textbf{Method} & \textbf{Smearing matrix} \\
        \toprule
        Standard & $\Identity$ \\
        Label smoothing & $(1 - \alpha) \cdot \Identity + \frac{\alpha}{L} \cdot \Ones$ \\
        Backward correction & $\frac{1}{1 - \rhoPrime} \cdot \Identity - \frac{\rhoPrime}{(1 - \rhoPrime) \cdot L} \cdot \Ones$ \\
        \bottomrule
    \end{tabular}
    \caption{Comparison of different label smearing methods.
    Here, 
    $\Identity$ denotes the identity and
    $\Ones$ the all-ones matrix.
    For backward correction, the theoretical optimal choice of $\alpha = \frac{L}{L - 1} \cdot \rho$,
    where $\rho$ is the level of symmetric label noise.}
    \label{tbl:comparison}
\end{table}

%
\subsection{Statistical consistency of label smearing}

Recall that our fundamental goal is to devise a procedure that can approximately minimise the population risk $R( \mathbf{f} )$.
Given this,
it behooves us to understand the effect of label smearing on this risk.
As we shall explicate, label smearing:
\begin{enumerate}[label=(\roman*),itemsep=0pt,topsep=0pt]
    \item is equivalent to fitting to a modified distribution.
    \item preserves classification consistency for suitable $\SmearMatrix$.
\end{enumerate}

%

For (i),
observe that the
smeared loss has corresponding risk
\begin{align*}
    R_{\mathrm{sm}}( \mathbf{f} ) 
    &= \E{\X}{ \mathbf{p}^{*}( \X )^{\tup} \ell^{\mathrm{SM}}( \mathbf{f}( \X ) ) } \\
    &= \E{\X}{ \mathbf{p}^{*}( \X )^{\tup} \SmearMatrix \, \ell( \mathbf{f}( \X ) ) }.
\end{align*}
Consequently, 
minimising a smeared loss
is equivalent to minimising the original loss 
on a \emph{smeared} distribution with class-probabilities $\pSmearStar( x ) = \SmearMatrix^{\tup} \pStar( x )$.

For example,
under label smoothing,
we fit to the class-probabilities
$\SmearMatrix^{\tup} \mathbf{p}^*( x ) = (1 - \alpha) \cdot \mathbf{p}^*( x ) + \frac{\alpha}{L}$.
This corresponds to a scaling and translation of the original.
This trivially preserves the label with maximal probability, provided $\alpha < 1$.
Smoothing
is 
thus \emph{consistent} for classification,
i.e.,
minimising its risk also minimises the classification risk~\citep{Zhang:2004,Zhang:2004b,Bartlett:2006}.

Now consider backward correction with $\mathbf{M} = \NoiseMat^{-1}$.
Suppose this is applied to a distribution with class-conditional label noise 
governed by transition matrix $\NoiseMat$.
Then,
we will fit to probabilities
$\SmearMatrix^{\tup} \bar{\mathbf{p}}^*( x ) = ( \NoiseMat^{\tup} )^{-1} \bar{\mathbf{p}}^*( x )$.
By~\eqref{eqn:noisy-eta-ccn},
these will exactly equal the \emph{clean} probabilities $\mathbf{p}^*( x )$;
i.e.,
backward correction will effectively denoise the labels.



%
\begin{figure*}[!t]
    \centering
    \subfigure[Label smoothing.]{%
    \includegraphics[scale=0.6]{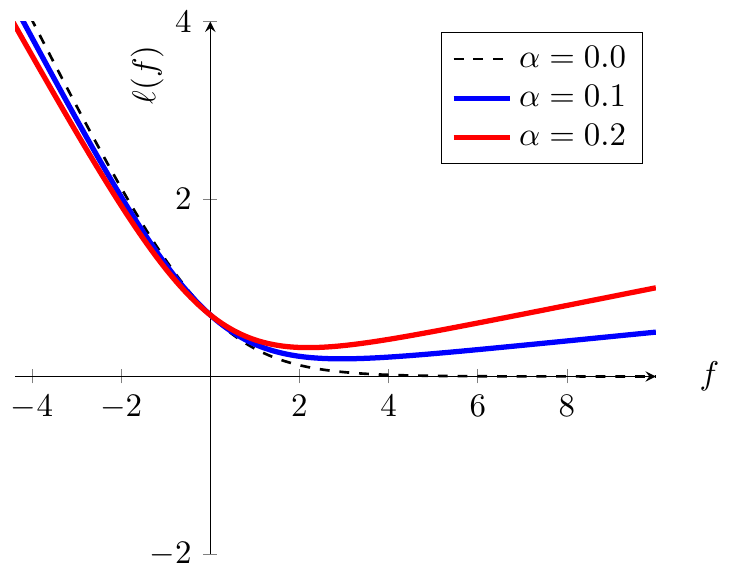}%
    }
    \subfigure[Backward correction.]{%
    \includegraphics[scale=0.6]{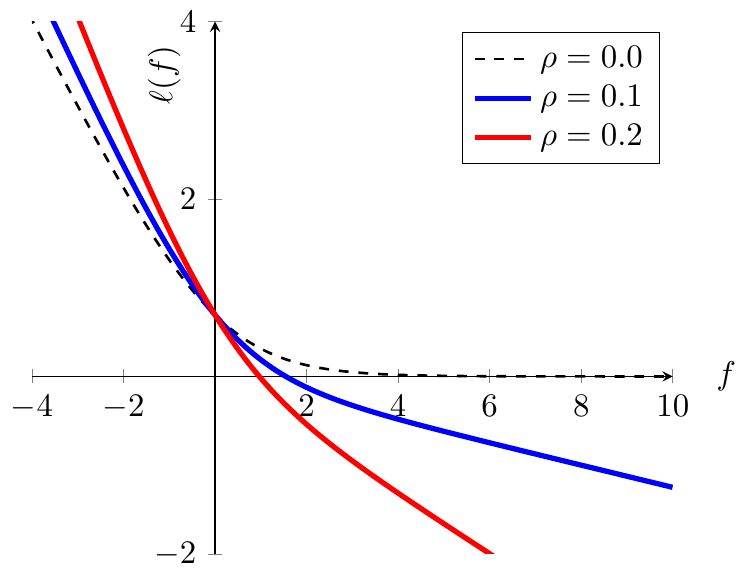}%
    }
    \subfigure[Forward correction.]{%
    \includegraphics[scale=0.6]{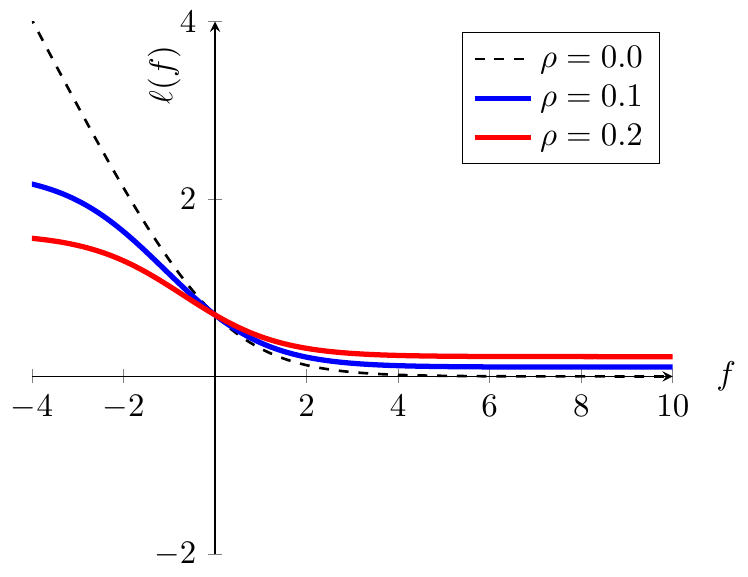}%
    }    
    \caption{Effect of label smoothing, backward correction, and forward correction on the logistic loss.
    The standard logistic loss vanishes for large positive predictions,
    and is linear for large negative predictions.
    Smoothing introduces a finite positive minima.
    Backward correction makes the loss \emph{negative} for large positive predictions.
    Forward correction makes the loss \emph{saturate} for large negative predictions.
    }
    \label{fig:smooth_logistic}
\end{figure*}




%
\subsection{How does label smoothing relate to loss correction?}
\label{sec:smoothing_vs_smearing}

Following
Table~\ref{tbl:comparison},
one cannot help but notice a strong similarity between label smoothing 
and backward correction for symmetric noise.
Both methods combine an identity matrix with an all-ones matrix;
the striking difference, however,
is that this combination is via \emph{addition} in one, but \emph{subtraction} in the other.
This results in losses with very different forms:
\begin{align*}
    \numberthis
    \label{eqn:smoothing-vs-backward}
    \ell^{\mathrm{LS}}( y, \mathbf{f} ) &\propto \ell( y, \mathbf{f} ) + \frac{\alpha}{(1 - \alpha) \cdot L} \cdot \sum_{y'} \ell( y', \mathbf{f} ) \\
    \ellBack( y, \mathbf{f} ) &\propto \ell( y, \mathbf{f} ) - \frac{\rhoPrime}{L} \sum_{y'} \ell( y', \mathbf{f} ).    
\end{align*}
Fundamentally, the effect of the two techniques is different:
smoothing aims to \emph{minimise}
the average per-class loss
$\frac{1}{L} \sum_{y'} \ell( y', \mathbf{f} )$,
while backward correction seeks to \emph{maximise} this.
Figure~\ref{fig:smooth_logistic} visualises the effect on the losses when $L = 2$,
and $\ell$ is the logistic loss.
Intriguingly, the smoothed loss is seen to penalise confident predictions.
On the other hand, backward correction allows one to compensate for overly confident negative predictions by allowing for a \emph{negative} loss on positive samples that are correctly predicted.


Label smoothing also relates to {forward correction}:
recall that here, we compute the loss
$ \ell^{\rightarrow}( \mathbf{f} ) = \ell( \mathbf{T} \mathbf{f} ). $
Compared to label smoothing,
forward correction thus performs smoothing of the \emph{logits}.
As shown in Figure~\ref{fig:smooth_logistic},
the effect is that the loss becomes bounded for all predictions.

At this stage,
we return to our original motivating question:
can label smoothing mitigate label noise?
The above would seem to indicate otherwise:
backward correction guarantees an unbiased risk estimate,
and yet we have seen smoothing constructs a fundamentally different loss.
In the next section, we assess whether this is borne out empirically.

%% file: experiments_denoise.tex
We now present experimental observations of the effects of label smoothing under label noise.
We then provide insights into why smoothing can successfully denoise labels,
by viewing smoothing as a form of \emph{shrinkage regularisation}.

%
\subsection{Denoising effects of label smoothing}

We begin by empirically answering the question: can label smoothing successfully mitigate label noise?
To study this,
we employ smoothing in settings where the training data is artificially injected with symmetric label noise.
This follows the convention in the label noise literature~\citep{Patrini:2017,Han:2018b,Charoenphakdee:2019}.

Specifically, 
we consider the \cifarT{}, \cifarH{} and \imagenet{} datasets, 
and add symmetric label noise at level $\rho^* = 20\%$ to the training (but \emph{not} the test) set;
i.e.,
we replace the training label with a uniformly chosen label $20\%$ of the time.
On \cifarT{} and \cifarH{} we train two different models on this noisy data, \resnetT{} and \resnetF{},
with similar hyperparameters as~\citet{Muller:2019}.
Each experiment is repeated five times, and we report the mean and standard deviation of the \emph{clean} test accuracy.
On \imagenet{} we train \resnetIM{} with LARS~\cite{you2017large}.
We describe the detailed experimental setup in Appendix~\ref{sec:appx_hyper}.

As loss functions, 
our baseline
is
training with the softmax cross-entropy on the noisy labels.
We then employ label smoothing~\eqref{eqn:smoothing} ({\bf LS}) for various values of $\alpha$,
as well as forward ({\bf FC}) and backward ({\bf BC}) correction~\eqref{eqn:backward-correction},~\eqref{eqn:forward-correction}  assuming symmetric noise for various values of $\alpha$.
We remark here that in the label noise literature, it is customary to \emph{estimate} $\alpha$,
with theoretical optimal value $\alpha^* = \frac{L}{L - 1} \cdot \rho^*$;
however, we shall here simply treat this as a tuning parameter akin to the smoothing $\alpha$, 
whose effect we shall study.

We now analyse the results along several dimensions.

\begin{figure*}[!t]%
    \centering
    \subfigure[CIFAR 100]{%
    \label{fig:first}%
    \includegraphics[height=1.75in]{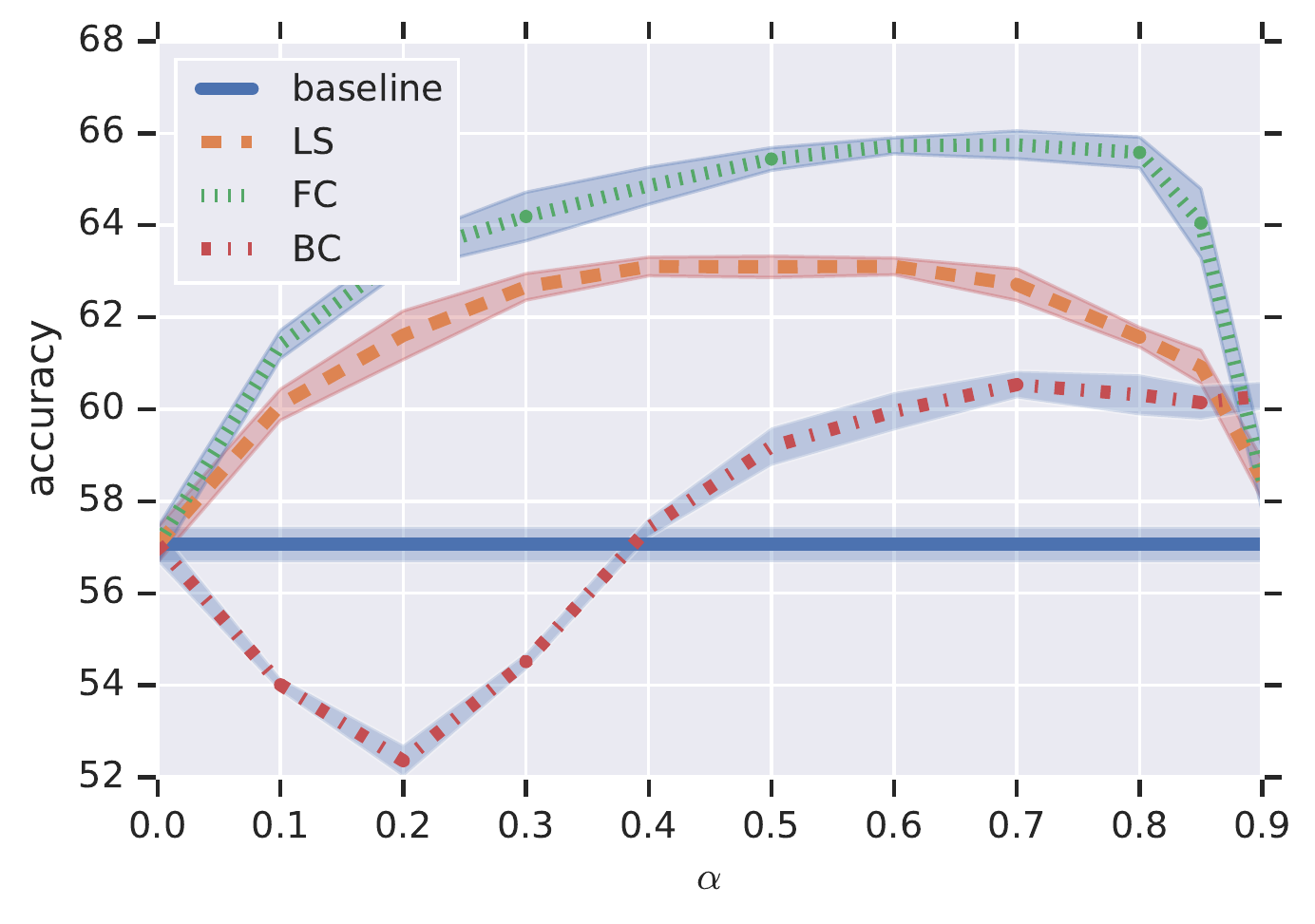}}%
    \qquad
    \subfigure[CIFAR 10]{%
    \label{fig:second}%
    \includegraphics[height=1.75in]{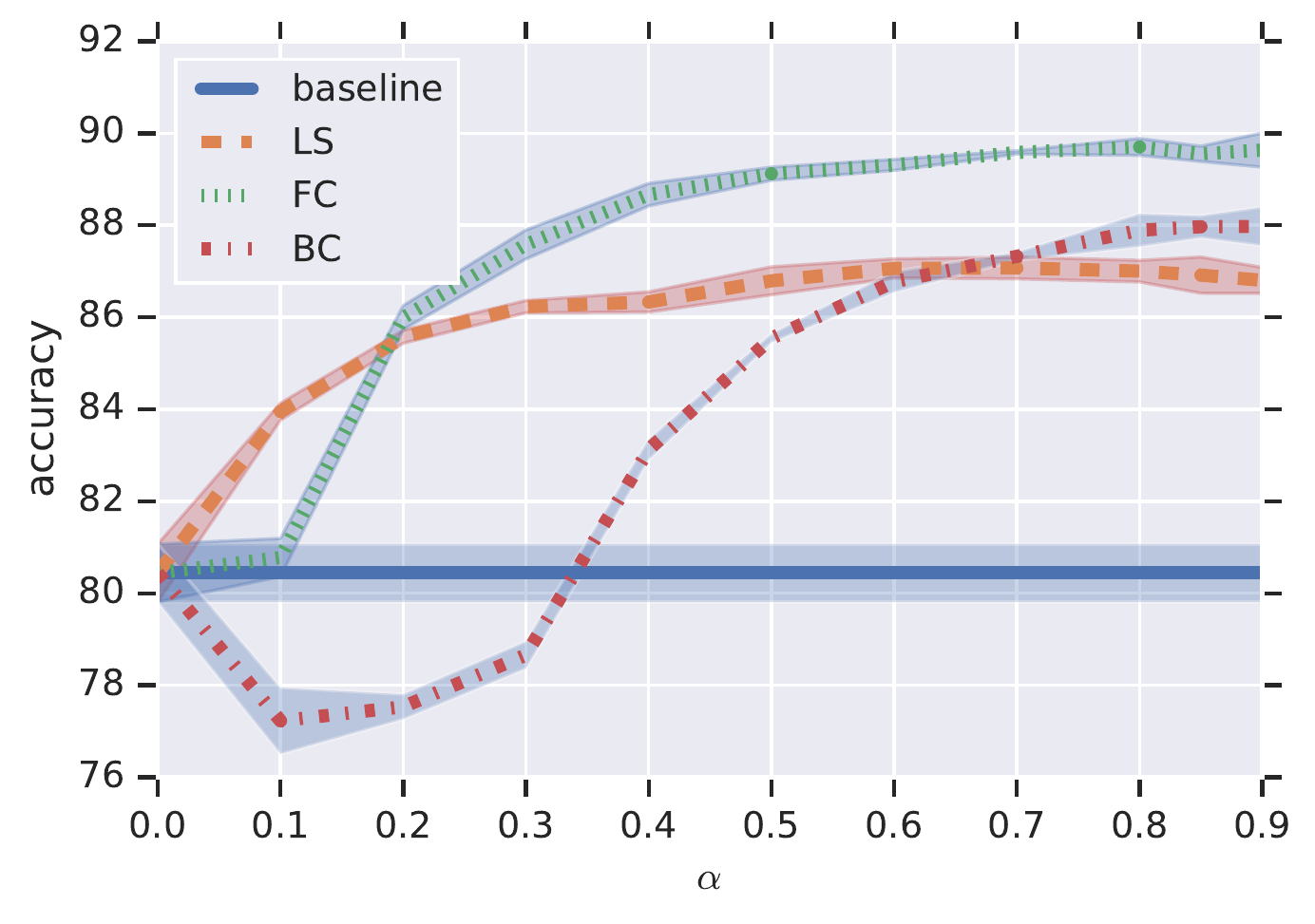}}%
    \caption{Effect of $\alpha$ on smoothing and forward label correction test accuracies on \cifarH{} and \cifarT{} from \resnetT{}. Standard deviations are denoted by the shaded regions. Label smoothing (LS) significantly improves over baseline, and choosing $\alpha \gg \rho^*$, the true noise rate, improves even further. Forward correction (FC) outperforms LS and also benefits from choosing large values for $\alpha$. Backward correction (BC) is worse than baseline for small $\alpha$, and better than baseline for large $\alpha$. In Table~\ref{tbl:experiments_smearing} in appendix, we report additional results for \resnetF{} and \resnetT{} from different label smearing methods, including where confusion matrix is estimated by pre-training a model as in ~\citet{Patrini:2017}.}
    \label{fig:cifar-smoothing-vs-correction}
\end{figure*}

\noindent {\bf Accuracy}:
In Figure \ref{fig:cifar-smoothing-vs-correction}, we plot the test accuracies of all methods on \cifarT{} and \cifarH{}.
Our first finding is that
\emph{label smoothing significantly improves accuracy over the baseline}.
We observe similar denoising effects on \imagenet{} in Table~\ref{tbl:imagenet}.
This confirms that empirically, label smoothing is effective in dealing with label noise. 

Our second finding is that, surprisingly,
\emph{choosing $\alpha \gg \rho^*$, the true noise rate, improves performance of all methods}.
This is in contrast to the theoretically optimal choice $\alpha \approx \rho^*$
for loss correction approaches~\citep{Patrini:2017},
and indicates it is valuable to treat $\alpha$ as a tuning parameter.

Finally, we see that label smoothing is \emph{often competitive with loss correction}.
This is despite it minimising a fundamentally different loss to the unbiased backward correction,
as discussed in~\S\ref{sec:smoothing_vs_smearing}.
We note however that loss correction generally produces the best overall accuracy with high $\alpha$.

\begin{table}[!t]
    \centering
    
    {
    \begin{tabular}{lccccc@{}}
        \toprule
        \multicolumn{1}{c}{\textbf{}} & \textbf{$\alpha = 0.0$} & \textbf{$\alpha = 0.1$} & \multicolumn{1}{c}{$\alpha = 0.2$} &  \multicolumn{1}{c}{$\alpha = 0.4$} &\multicolumn{1}{c}{$\alpha = 0.6$}\\
        \toprule
        LS & 70.86 & 71.12 & 71.55 & 70.95 & 70.59\\ 
        FC & 70.86 & 73.04 & 73.17 & 73.35 & 72.92\\ 
        \bottomrule
    \end{tabular}
    }
    
    \caption{Test accuracy on \imagenet{} trained with 
    $\rho=20\%$
    label noise on \resnetIM{}, with
    label smoothing (LS) and forward correction (FC)
    for varying $\alpha$. Both LS and FC successfully denoise, and thus improve over the baseline ($\alpha = 0$).}
    \label{tbl:imagenet}
\end{table}

\begin{figure*}[!t]
    \centering
    \subfigure[Gap between true (unobserved) label logit and mean logit.]{%
    \includegraphics[width=0.35\textwidth]{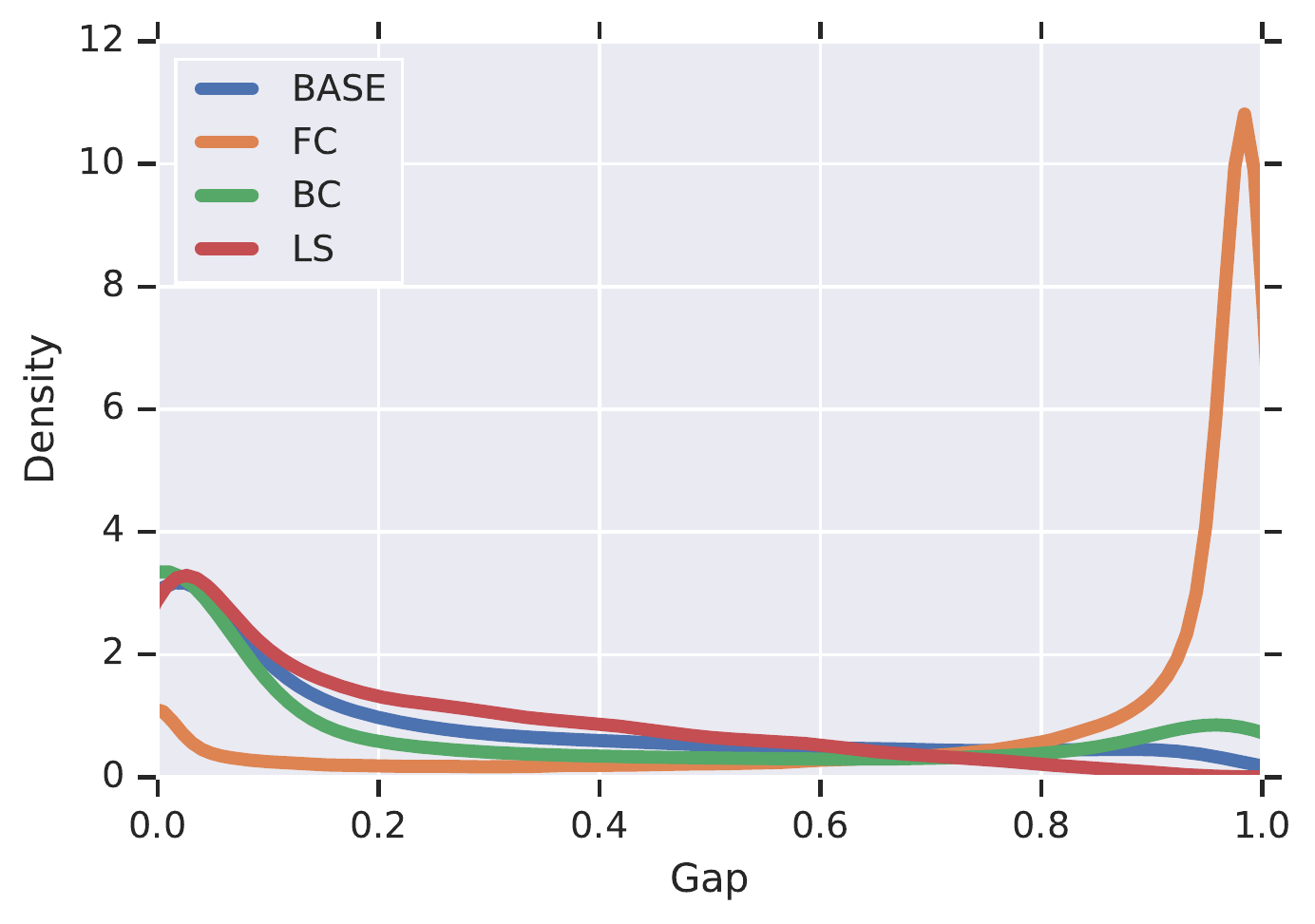}%
    \label{fig:noise_labellogit_vs_avglogit_true}
    }%
    \qquad
    \subfigure[Gap between noisy (observed) label logit and mean logit.]{%
    \includegraphics[width=0.35\textwidth]{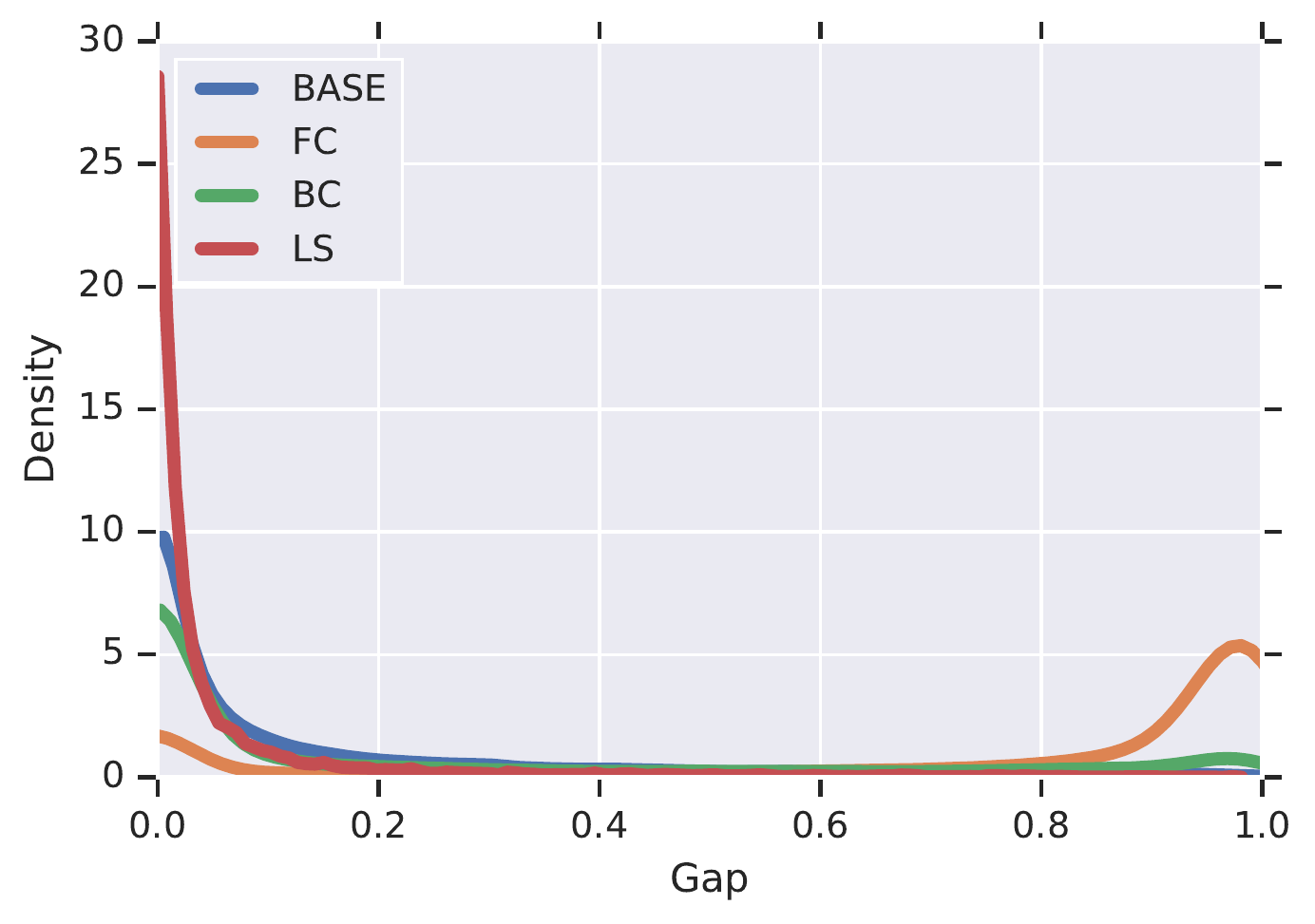}%
    \label{fig:noise_labellogit_vs_avglogit_noisy}
    }    
    \caption{Density of differences between logit corresponding to the 
    true (left plot; corresponding to the ``true'' label, before injecting label noise) and 
    noisy label (right plot; corresponding to the ``noisy'' label, after injecting label noise) 
    and the average over all logits on the mis-labeled portion of the train data.
    Results are with $\alpha=0.2$ on \cifarH{}, and the \resnetT{} model.
    LS reduces confidence mostly on the noisy label, whereas FC and BC increase confidence mostly on the true label.
    See Figure~\ref{fig:maxlogit_vs_avglogit} for plots on full and clean data.}
    \label{fig:noise_labellogit_vs_avglogit}
\end{figure*}

\noindent {\bf Denoising}:
What explains the effectiveness of label smoothing for training with label noise? 
{Does it 
correct the predictions on noisy examples,
or does it 
only further improve the predictions 
on the clean (non-noisy) examples?}

To answer these questions, we separately inspect accuracies on the noisy and clean portions of the training data (i.e., on those samples whose labels are flipped, or not).
Table~\ref{tbl:acc_on_different_portions_of_train} reports this breakdown from the \resnetT{} model on \cifarH{}, for different values of $\alpha$. 
We see that as $\alpha$ increases, accuracy improves on both the noisy and clean parts of the data, 
with a more significant boost on the noisy part.
Consequently, smoothing \emph{systematically improves predictions} of both clean and noisy samples.

%
\noindent {\bf Model confidence}:
Predictive accuracy is only concerned with a model ranking the true label ahead of the others.
However, the \emph{confidence} in model predictions is also of interest,
particularly since a danger with label noise is 
being overly confident in predicting a noisy label.
How do smoothing and correction methods affect this confidence under noise?

To measure this,
in Figure~\ref{fig:noise_labellogit_vs_avglogit} 
we plot distributions of the differences between the logit activation $\hat{p}( y \mid x )$ for a true/noisy label $y$, 
and the average logit activation $\frac{1}{L} \sum_{y' \in [L]} \hat{p}( y' \mid x )$. 
Compared to the baseline, label smoothing significantly \emph{reduces confidence} in the \emph{noisy} label 
(refer to the left side of Figure~\ref{fig:noise_labellogit_vs_avglogit_noisy}).

\begin{figure*}[!t]
    \centering
    \subfigure[Label smoothing $\alpha=0$.]{%
    \includegraphics[width=0.275\textwidth]{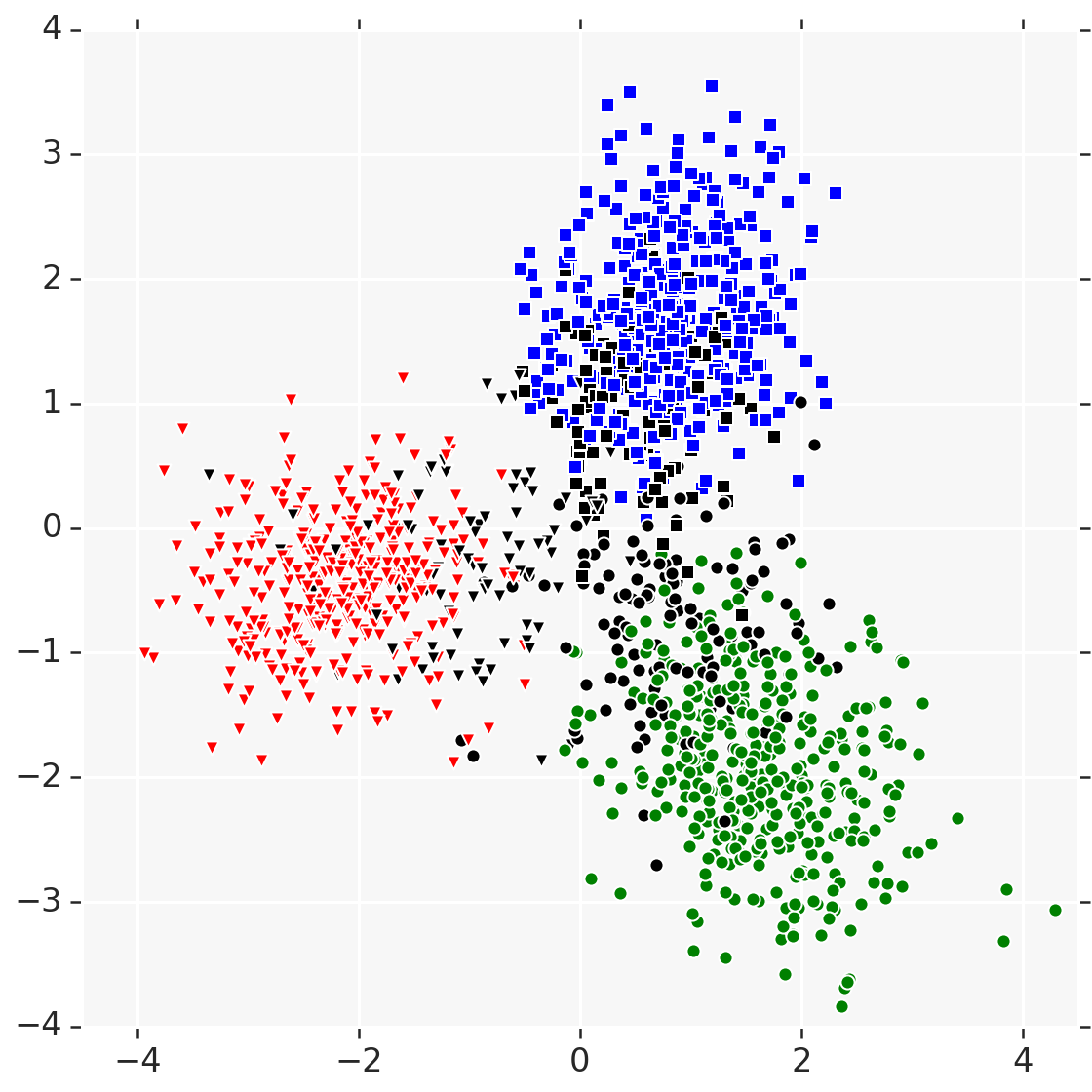}%
    }
    \subfigure[Label smoothing $\alpha=0.2$.]{%
    \includegraphics[width=0.275\textwidth]{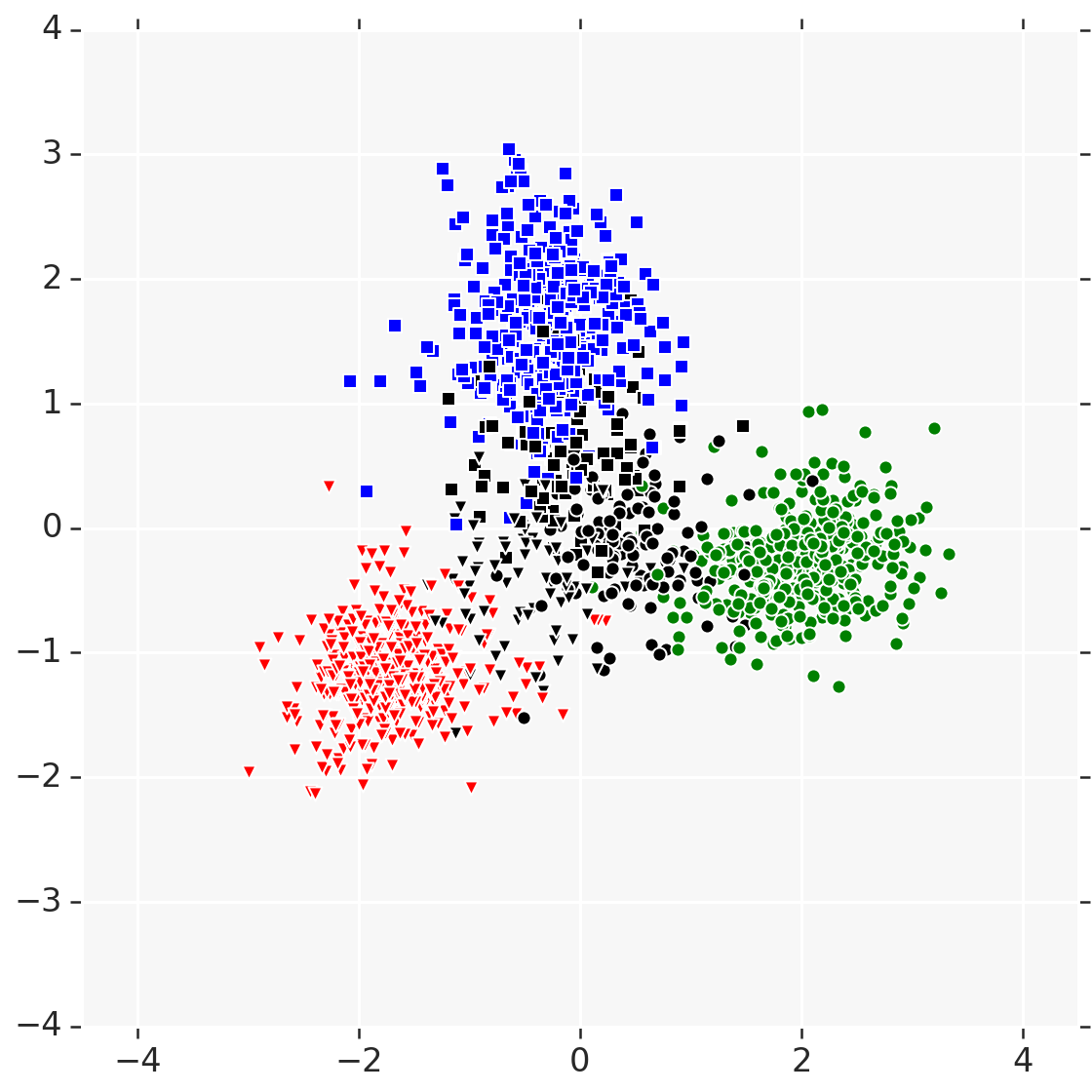}%
    }
    \subfigure[Label smoothing $\alpha=0.7$.]{%
    \includegraphics[width=0.275\textwidth]{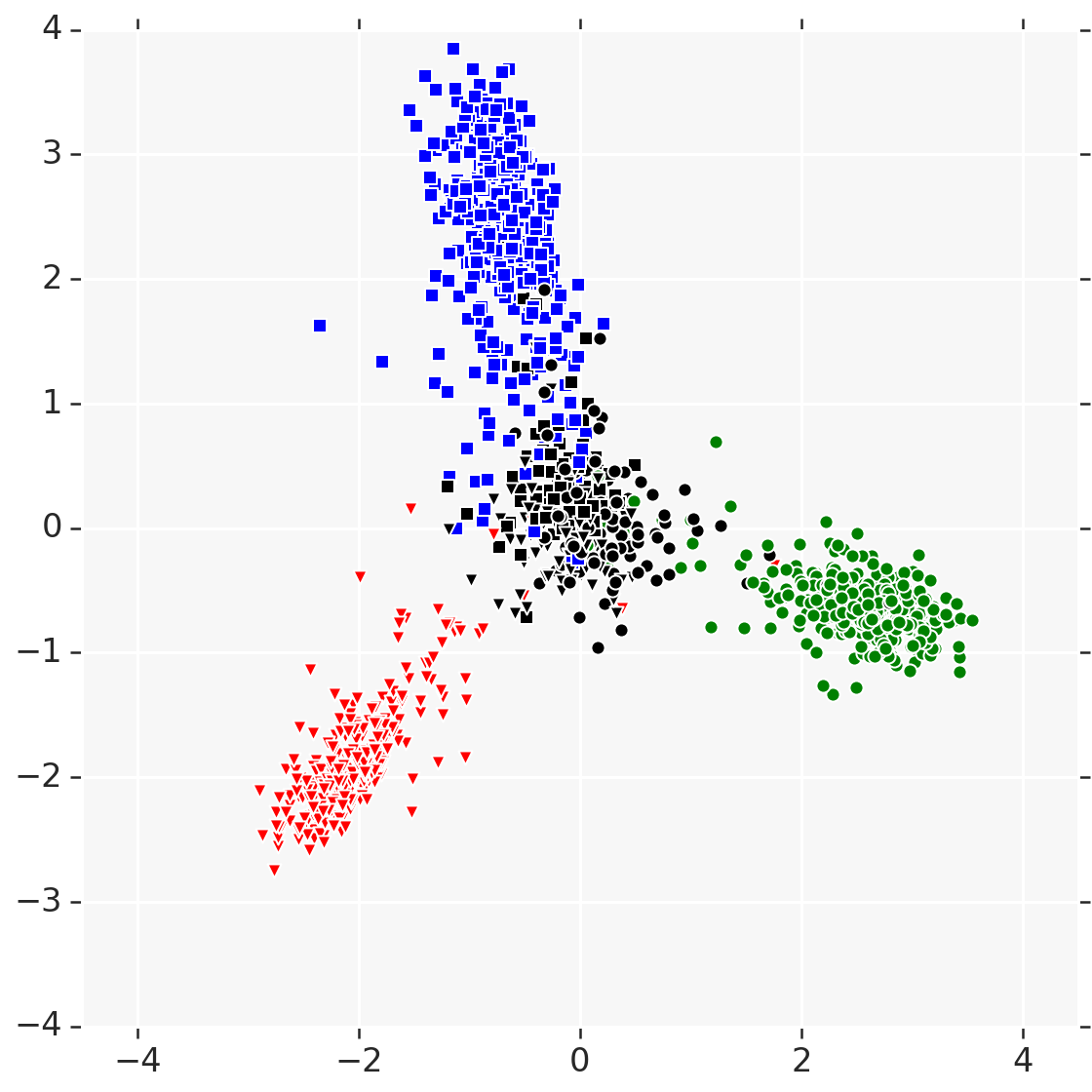}%
    }    
    \caption{Effect of label smoothing on pre-logits (penultimate layer output) under label noise.
    Here, we visualise the pre-logits of a \resnetT{} for three classes on \cifarH{}, using the procedure of~\citet{Muller:2019}.
    The black markers denote instances which have been labeled incorrectly due to noise.
    Smoothing is seen to have a denoising effect: the noisy instances' pre-logits become more uniform, and so the model learns to not be overly confident in their label.}
    \label{fig:smoothing_logits}
\end{figure*}

To visualise this effect of smoothing, 
in Figure~\ref{fig:smoothing_logits} we plot pre-logits (penultimate layer output) of examples from 3 classes projected onto their class vectors as in~\citet{Muller:2019}, for a \resnetT{} trained on \cifarH{}.
As we increase $\alpha$, the confidences for noisy labels shrink, showing the denoising effects of label smoothing.

On the other hand, both forward and backward correction systematically \emph{increase confidence} in predictions.
This is especially pronounced for forward correction, demonstrated by the large spike for high differences in Figure~\ref{fig:noise_labellogit_vs_avglogit_noisy}.
At the same time, these techniques increase the confidence in predictions of the true label 
(refer to Figure~\ref{fig:noise_labellogit_vs_avglogit_true}):
forward correction in particular
becomes much more confident in the true label than any other technique.

In sum,
Figure~\ref{fig:noise_labellogit_vs_avglogit} illustrates both positive and adverse effects on confidence from label smearing techniques:
label smoothing becomes less confident in both the noisy and correct labels,
while forward and backward correction become more confident in both the correct labels and noisy labels.

\begin{table}[!t]
    \centering
    
    {
    \begin{tabular}{@{}lcccc@{}}
        \toprule
        \multicolumn{1}{c}{\textbf{$\alpha$}} & \multicolumn{1}{c}{\textbf{Full train}} & \textbf{Clean part of train} & \multicolumn{2}{c}{\textbf{Noisy part of train}}\\
         & true labels & true labels & true labels & noisy labels\\
        
        \toprule
        0.0 & 77.39 & 86.75 & 39.92 & 17.88 \\ 
        0.1 & 80.11 & 87.99 & 48.58 & 12.27\\
        0.2 & 81.22 & 88.27 & 53.01 & 8.32\\
        \bottomrule
    \end{tabular}
    }
    
\caption{Accuracy on different portions of the training set from \resnetT{}, trained with different label smoothing values $\alpha$ on \cifarH{}. As $\alpha$ increases, accuracy improves on both clean and noisy part of data. Interestingly, the improvement on the noisy part of data is greater than the reduction in fit to the noisy labels (compare the two rightmost columns in the table). Thus, there are noisy examples assigned neither to correct class nor to the observed \emph{noisy} class without LS, and which LS helps classify correctly.}
    \label{tbl:acc_on_different_portions_of_train}
\end{table}

\begin{table}[!t]
    \centering
    
    
    {
    \begin{tabular}{@{}lccc@{}}
        \toprule
        \multicolumn{1}{c}{\textbf{$\alpha$}} & \textbf{LS} & \multicolumn{1}{c}{\textbf{FC}} & \multicolumn{1}{c}{\textbf{BC}}\\
        \toprule
        0.0 & 0.111 & 0.111 & 0.111 \\ 
        0.1 & 0.108 & 0.153 & 0.214 \\ 
        0.2 & 0.156 & 0.165 & 0.266 \\ 
        \bottomrule
    \end{tabular}
    }
    
    \caption{Expected calibration error (ECE) computed on $100$ bins on test set
    for \resnetT{} on \cifarH{}, trained with different label smearing techniques under varying values of $\alpha$.
    Generally, label smearing is detrimental to calibration.}
    \label{tbl:ece}
\end{table}

\noindent {\bf Model calibration}:
To further tease out the impact of label smearing on model confidences,
we ask:
how do these techniques affect the \emph{calibration} of the output probabilities?
This measures how meaningful the model probabilities are in a frequentist sense~\citep{Dawid:1982}.

In Table~\ref{tbl:ece}, we report
the expected calibration error (ECE)~\citep{Guo:2017}
on the test set for each method. 
While smoothing improves calibration over the baseline with $\alpha=0.1$
--- an effect noted also in~\cite{Muller:2019} ---
for larger $\alpha$, it becomes significantly \emph{worse}.
Furthermore, loss correction techniques significantly degrade calibration over smoothing.
This is in keeping with the above findings as to these methods sharpening prediction confidences.

\begin{figure*}[!t]
    \centering
    \subfigure[Label smoothing.]{%
    \includegraphics[width=0.35\textwidth]{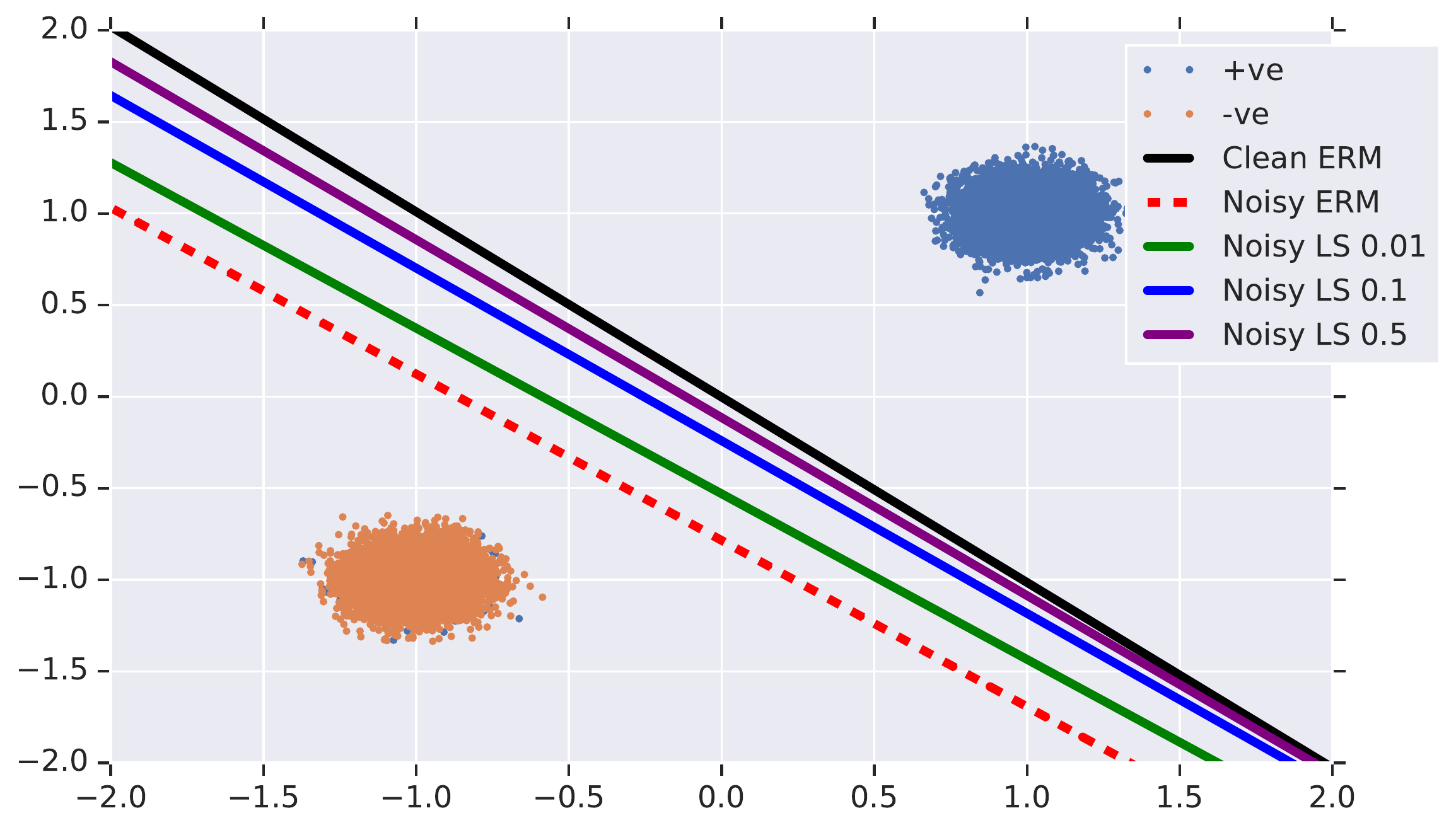}%
    }
    \subfigure[$\ell_2$ regulariser.]{%
    \includegraphics[width=0.35\textwidth]{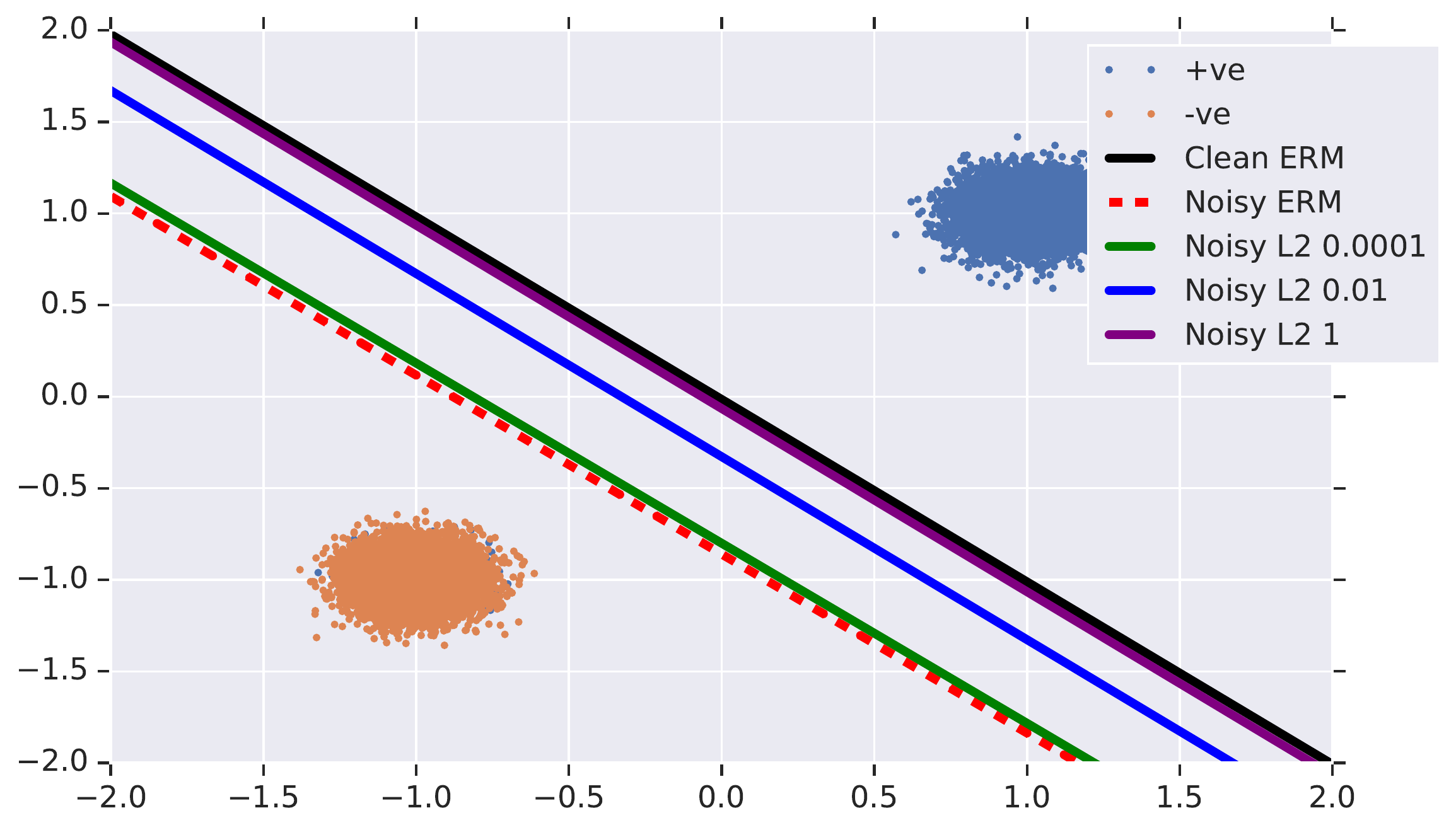}%
    }
    \caption{(a) Effect of label smoothing on logistic regression separator,
    on a synthetic problem with asymmetric label noise.
    The black line is the Bayes-optimal separator, found by logistic regression on the clean data.
    The other lines are separators learned by applying label smoothing with various $\alpha$ on the noisy data.
    Without smoothing, noise draws the separator towards the affected class;
    smoothing undoes this effect, and brings the separator back to the Bayes-optimal. 
    (b) Shrinkage ($\ell_2$) regularisation has a similar effect on the separator.}
    \label{fig:synth_asymmetric_noise}
\end{figure*}

\noindent \textbf{Summary}:
Overall, our results demonstrate that 
label smoothing is competitive with loss correction techniques in coping with label noise,
and that it is particularly successful in denoising examples
while preserving calibration.

%% file: smoothing_regularisation.tex
\subsection{Label smoothing as regularisation}

While empirically encouraging, the results in the previous section
indicate a gap in our theoretical understanding:
from
\S\ref{sec:smoothing_vs_smearing}, the smoothing loss apparently has the \emph{opposite} effect to backward correction, which is theoretically unbiased under noise.
What, then, explains the success of smoothing?

To understand the denoising effects of label smoothing,
we now study its role as a \emph{regulariser}.
{To get some intuition, consider a linear model $\mathbf{f}( x ) = \mathbf{W} x$, 
trained on features 
$\mathbf{X} \in \mathbb{R}^{N \times D}$ and 
one-hot labels $\mathbf{Y} \in \{ 0, 1 \}^{N \times L}$ using the square loss,
i.e.,
$\min_{\mathbf{W}} \| \mathbf{X} \mathbf{W} - \mathbf{Y} \|_2^{2}$.
Label smoothing at level $\alpha$ transforms the optimal solution 
$\mathbf{W}^*$ to
\begin{equation}
    \label{eqn:opt-smoothing-linreg}
    \bar{\mathbf{W}}^* = (1 - \alpha) \cdot \mathbf{W}^* + \frac{\alpha}{L} \cdot (\mathbf{X}^{\tup} \mathbf{X})^{-1} \mathbf{X}^{\tup} \Ones.
\end{equation}
Observe that if our data is centered, the second term will be zero.
Consequently,
for such data,
{the effect of label smoothing is simply to shrink the weights}.
Thus, \emph{label smoothing can have a similar effect to shrinkage regularisation}.}

%
Our more general finding is the following.
From~\eqref{eqn:smoothing-vs-backward}, label smoothing is equivalent to minimising a \emph{regularised} risk
$R_{\mathrm{sm}}( \mathbf{f}; D ) \propto R( \mathbf{f}; D ) + \beta \cdot \Omega( \mathbf{f} )$,
where
\begin{align*}
    \Omega( \mathbf{f} ) &\defEq \E{\X}{ \sum_{y' \in [L]} \ell( y', \mathbf{f}( \X ) ) },
\end{align*}
and $\beta \defEq \frac{\alpha}{(1 - \alpha) \cdot L}$.
The second term above does \emph{not} depend on the underlying label distribution $\Pr( \Y \mid \X )$.
Consequently, it may be seen as a \emph{data-dependent regulariser} on our predictor $\mathbf{f}$.
Concretely, for the softmax cross-entropy,
\begin{align}
\Omega( \mathbf{f} ) = \E{\X}{ L \cdot \log\left[ \sum_{y'} e^{f_{y'}( \X )} \right] - \sum_{y'} f_{y'}( \X ) }. \label{eq:ls_regulariser}
\end{align}

To understand the label smoothing regulariser~\eqref{eq:ls_regulariser} more closely,
we study it for the special case of linear classifiers, i.e., $f_{y'}( \X ) = \inp{\w_{y'}}{\X}.$ 
While we acknowledge that the label smoothing effects displayed in our experiments for deep networks are complex, 
as a first step, 
understanding these effects for simpler models will prove instructive.

%
\textbf{Smoothing for linear models}.
For linear models $f_{y'}( \X ) = \inp{\w_{y'}}{\X}$,
the label smoothing regularization for softmax cross-entropy~\eqref{eq:ls_regulariser} 
induces the following \emph{shrinkage} effect.

\begin{theorem}\label{thm:ls}
Let $\X$ be distributed as $\mathbb{Q}$ with a finite mean. Then $\w_{y'}=0, \forall y' \in [L]$ is the minimiser of \eqref{eq:ls_regulariser}.
\end{theorem}

See Appendix~\ref{sec:proofs} for the proof.
We see that 
the label smoothing regulariser encourages shrinkage of our weights towards zero;
this is akin to the observation for square loss in~\eqref{eqn:opt-smoothing-linreg},
and similar in effect to $\ell_2$ regularisation, which is also motivated as increasing the classification margin.

This perspective gives one hint as to why smoothing may successfully denoise.
For linear models, introducing asymmetric label noise can move the decision boundary closer to a class.
Hence, a regulariser that increases margin, 
such as shrinkage,
can help the model to be more robust to noisy labels.
We illustrate this effect with
the following experiment. 

%
\textbf{Effect of shrinkage on label noise}.
We consider a 2D problem comprising Gaussian class-conditionals,
centered at $\pm ( 1, 1 )$
and with isotropic covariance at scale $\sigma^2 = 0.01$.
The optimal linear separator is one that passes through the origin, shown in Figure~\ref{fig:synth_asymmetric_noise} as a black line.
This separator is readily found by fitting logistic regression on this data.

\begin{table*}[!t]
    \centering
    
    \resizebox{0.99\linewidth}{!}{
    \begin{tabular}{@{}lllllll@{}}
        \toprule
        \textbf{Dataset} &\textbf{Architecture} & \textbf{Vanilla distillation} & \textbf{LS on teacher}  & \textbf{LS on student} & \textbf{FC on teacher} & \textbf{FC on student}\\
        \toprule
        \cifarH & \resnetT & 63.98$\pm$0.26 & 64.48$\pm$0.25 & 63.83$\pm$0.28 & \best{66.65$\pm$0.18} & 63.94$\pm$0.34\\ 
        \cifarH & \resnetF & 64.31$\pm$0.26 & 65.63$\pm$0.24 & 64.50$\pm$0.32 & \best{66.35$\pm$0.20} & 64.24$\pm$0.26\\
        \midrule
        \cifarT & \resnetT & 80.44$\pm$0.64 & \best{86.95$\pm$1.82} & 85.72$\pm$2.61 & 86.81$\pm$1.86 & 86.92$\pm$2.11\\ 
        \cifarT & \resnetF & 77.98$\pm$0.25 & \best{87.10$\pm$1.66} & 86.98$\pm$1.71 & 86.88$\pm$1.80 & 86.82$\pm$1.76\\
        \bottomrule
    \end{tabular}
    }
    \caption{Knowledge distillation experiments. We use label smoothing parameter $\alpha=0.1$ and temperature parameter $T=2$ during distillation, for all these experiments. We notice that doing LS on teacher improves the student accuracy compared to the baseline. LS on the student helps as well but not to the same accuracy. Loss correction using FC on teacher helps as well with the distillation.}
    \label{tbl:experiments_distillation}
\end{table*}

We inject 5\% \emph{asymmetric} label noise into the negatives, so that some of these have their labels flipped to be positive.
The effect of this noise is to move the logistic regression separator closer to the (true) negatives, indicating there is greater uncertainty in its predictions.
However, 
if we apply label smoothing at various levels $\alpha$,
the separator is seen to gradually converge back to the Bayes-optimal;
this is in keeping with the shrinkage property of the regulariser~\eqref{eq:ls_regulariser}.

Further, 
as suggested by Theorem \ref{thm:ls},
an explicit $L_2$ regulariser has a similar effect to smoothing (Figure \ref{fig:synth_asymmetric_noise}(b)). 
Formally establishing the relationship between label smoothing and shrinkage 
is an interesting open question.

\textbf{Summary}.
We have seen in~\S\ref{sec:label_smearing} that from a \emph{loss} perspective,
label smoothing results in a biased risk estimate;
this is contrast to the unbiased backward correction procedure.
In this section, we provided an alternate \emph{regularisation} perspective,
which gives insight into why label smoothing can denoise training labels.
Combining these two views theoretically, however, remains an interesting topic for future work.

%



%% file: noisy_distillation.tex
We now study the effect of label smoothing on distillation, 
when our data is corrupted with label noise. 
In distillation, a trained ``teacher'' model's logits are used to augment (or replace) the one-hot labels used to train a ``student'' model \cite{Hinton:2015}.
While traditionally motivated as a means for a simpler model (student) to mimic the performance of a complex model (teacher),~\citet{Furlanello:2018} showed gains even for models of similar complexity.

\citet{Muller:2019} observed that
for standard (noise-free) problems,
label smoothing on the teacher 
\emph{improves} the teacher's performance,
but
\emph{hurts} the student's performance.
Thus, a better teacher 
does not result in a better student.
\citet{Muller:2019} attribute this to 
the erasure of relative information between the teacher logits under smoothing.

But is a teacher trained with label smoothing on \emph{noisy} data better for distillation? 
On the one hand, as we saw in previous section, label smoothing has a denoising effect on models trained on noisy data. 
On the other hand, label smoothing on clean data may cause some information erasure in logits~\citep{Muller:2019}.
Can the teacher transfer the denoising effects of label smoothing to a student?

We study this question empirically.
On the \cifarH{} and \cifarT{} datasets,
with the same architectures and noise injection procedure as the previous section,
we train three teacher models on the noisy labels:
one as-is on the noisy labels,
one with label smoothing, and another with forward correction.
We distill each teacher 
to a student model of the same complexity
(see Appendix~\ref{sec:appx_hyper} for a complete description),
and measure the student's performance.
As a final approach, we distill a vanilla teacher,
but apply label smoothing and forward correction on the \emph{student}.

Table~\ref{tbl:experiments_distillation} reports the performance 
of the distilled students
using each of the above teachers.
Our key finding is that 
on both datasets,
\emph{both label smoothing and loss correction on the teacher significantly improves over vanilla distillation};
this is in marked contrast to the findings of~\citet{Muller:2019}.
On the other hand, smoothing or correcting on the student has mixed results;
while there are benefits on \cifarT{},
the larger \cifarH{} sees essentially no gains.

\begin{figure}[!t]
    \centering
    \includegraphics[scale=0.425]{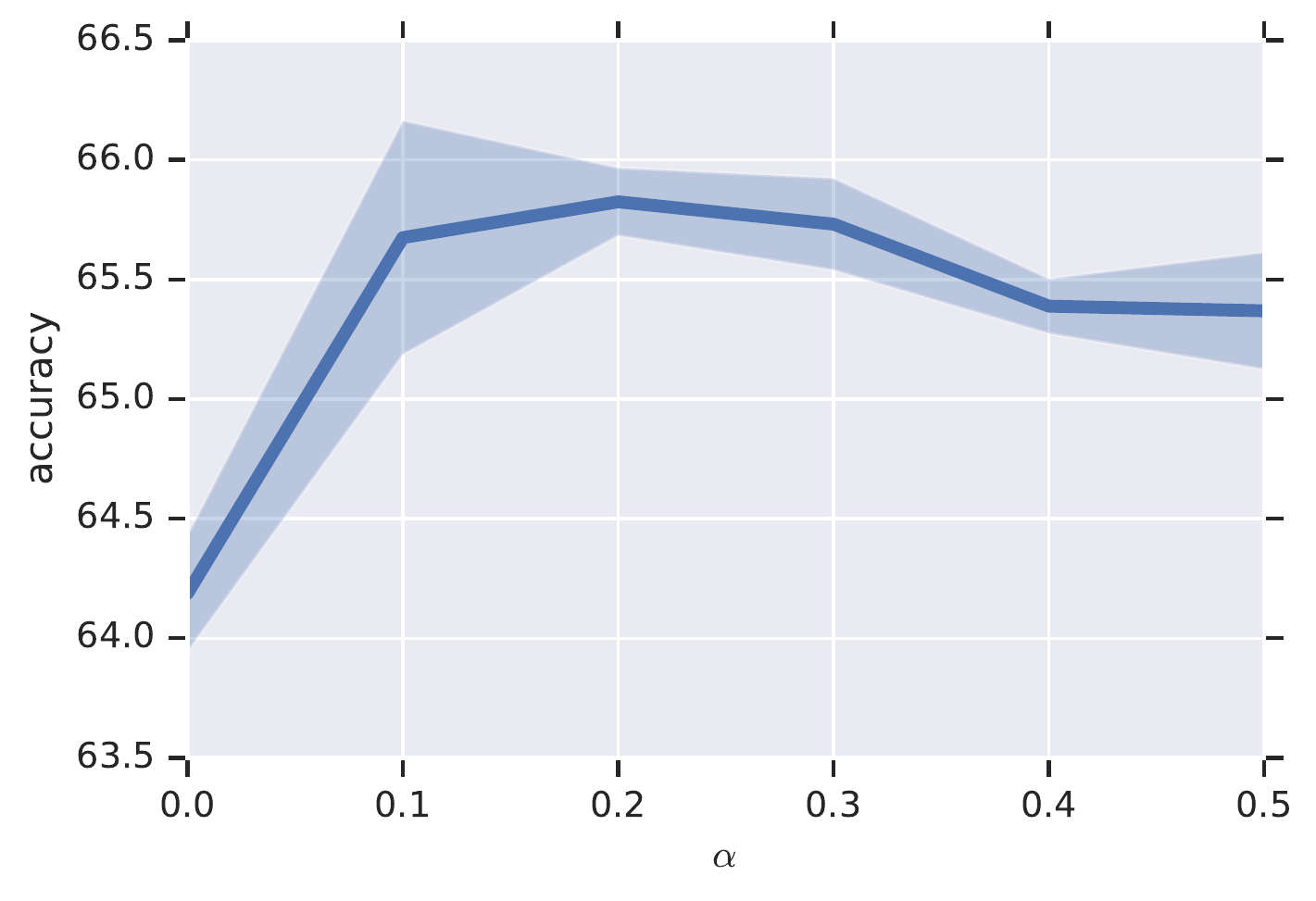}
    \caption{Effect of label smoothing on the teacher on student's accuracy after distillation with temperature $T=1$, \cifarH{}. 
    Teacher and student both use \resnetT{}. 
    For all values of $\alpha$, label smoothing on the teacher improves distillation performance compared to a plain teacher ($\alpha=0$).}
    \label{fig:sweep_alphas_for_temps_cifar100}
\end{figure}

Finally,
we plot the effect of the teacher's label smoothing parameter $\alpha$ on student performance in Figure~\ref{fig:sweep_alphas_for_temps_cifar100}.
Even for high values of $\alpha$, smoothing improves performance over the baseline ($\alpha=0$). 
Per the previous section, large values of $\alpha$ allow for successful label denoising,
and the results indicate the value of this transfer to the student.

In summary, our experiments show that under label noise, it is strongly beneficial to denoise the teacher 
---
either through label smoothing or loss correction
---
prior to distillation.

%

%% file: conclusion.tex
We studied the effectiveness of label smoothing as a means of coping with {label noise}.
Empirically, we showed that smoothing is competitive with existing loss correction techniques,
and that it exhibits strong denoising effects.
Theoretically, we related smoothing to one of these correction techniques,
and re-interpreted it as a form of regularisation.
Further, we showed that when distilling models from noisy data, 
label smoothing of the teacher is beneficial.
Overall, our results shed further light on the potential benefits of label smoothing,
and suggest formal exploration of its denoising properties as an interesting topic for future work.



%% file: appendix.tex
\begin{center}
  {\Large\bf Supplementary material for ``Does label smoothing mitigate label noise?''}
\end{center}

\section{Proof of Theorem~\ref{thm:ls}}
\label{sec:proofs}

%
Note that for linear models, $\Omega( \mathbf{f} )$ is a convex function of $\w_{y'}$. Hence  we can find the minimiser of $\Omega( \mathbf{f} )$ by solving for when the gradient vanishes. We have, 
\begin{align*} 
\frac{\partial \Omega( \mathbf{f} )}{\partial \w_{i}} &=  \E{\mathbb{Q}}{L \cdot \frac{e^{ \inp{\w_{i}}{ \X }}}{\sum_{y'} e^{ \inp{\w_{y'}}{ \X }}} \cdot \X - \X }. 
\end{align*} 
We can swap differential and expectation in the above equation as $\Omega( \mathbf{f} )$ is differentiable in both $\w_{i}$ and $\X$.
Now we show that the gradient evaluates to zero at $\w_i=0, \forall i$:
\begin{align*} 
\frac{\partial \Omega( \mathbf{f} )}{\partial \w_{i}} \bigg\rvert_{\w_i=0} &= \E{\mathbb{Q}}{L \cdot \frac{1}{\sum_{y'}1}\X -\X} \\
&=\E{\mathbb{Q}}{L \cdot \frac{1}{L} \X -\X} = 0.
\end{align*}

%
\section{Experimental setup}
\label{sec:appx_hyper}

\subsection{Architecture}
We use ResNet with batch norm \citep{he2016deep} for our experiments with the following configurations. For \cifarT{} and \cifarH{} we experiment with \resnetT{} and \resnetF{}. We use \resnetIM{} for our experiments with \imagenet{}. We list the architecture configurations in terms of ($\text{n}_\text{layer}$, $\text{n}_\text{filter}$, stride) corresponding to each ResNet block in Table \ref{tbl:architecture}.

\begin{table*}[!h]
    \centering
    
    \begin{tabular}{ll}
        \toprule
        \textbf{Architecture} &\textbf{Configuration:  [($\text{n}_\text{layer}$, $\text{n}_\text{filter}$, stride)]} \\
        \toprule
        \resnetT{} & [(5, 16, 1), (5, 32, 2), (5, 64, 2)] \\ 
        \resnetF{} & [(9, 16, 1), (9, 32, 2), (9, 64, 2)] \\
        \resnetIM{} & [(3, 64, 1), (4, 128, 2), (6, 256, 2), (3, 512, 2)]\\ 
        \bottomrule
    \end{tabular}
    \caption{ResNet Architecture configurations used in our experiments \citep{he2016deep}.}
    \label{tbl:architecture}
\end{table*}

\subsection{Training}

We follow the experimental setup from \citet{Muller:2019}. 
For both \cifarT{} and \cifarH{} we use a mini-batch size of $128$ and train for $64$k steps. We use stochastic gradient descent with Nesterov momentum of 0.9. We use an initial learning rate of $0.1$ with a schedule of dropping by a factor of $10$ at $32$k and $48$k steps.
We set weight decay to $\text{1e-4}$. On \imagenet{} we train \resnetIM{} using the LARS optimizer \cite{you2017large} for large batch training, with a batch size of $3500$, and training for $32768$ steps. For data augmentation we used random crops and left-right flips \footnote{\url{https://github.com/tensorflow/tensor2tensor/blob/master/tensor2tensor/data_generators/image_utils.py}}.

For our distillation experiments we train only with the cross-entropy objective against the teacher's logits.
We use a temperature of $2$ unless specified otherwise when describing an experiment.

We ran training on \cifarH{} and \cifarT{} using 4 chips of TPU v2 and \imagenet{} using 128 chips of TPU v3.
Training for \cifarH{} and \cifarT{} took under 15 minutes, and for \imagenet{} around $1.5$h.

\section{Experiments: additional results}

\subsection{Comparison of smoothing against label noise baselines}

\begin{table*}[!h]
    \centering
    
    \resizebox{0.99\linewidth}{!}{
    \begin{tabular}{llllllll}
        \toprule
        \textbf{Dataset} &\textbf{Architecture} & \textbf{Baseline} & \textbf{LS}  & \textbf{FC smoothing} & \textbf{BC smoothing} & \textbf{FC Patrini} & \textbf{BC Patrini}\\
        \toprule
        CIFAR100 & RESNET-32 & 57.06$\pm$0.38 & 60.70$\pm$0.28 & \best{61.29$\pm$0.38} & 53.91$\pm$0.40 & 57.25$\pm$0.24 & 55.89$\pm$0.33\\ 
        CIFAR100 & RESNET-56 & 54.93$\pm$0.37 & 59.04$\pm$0.53 & \best{60.00$\pm$0.31} & 52.25$\pm$0.51 & 55.09$\pm$0.39 & 55.00$\pm$0.13\\
        CIFAR10 & RESNET-32 & 80.44$\pm$0.63 & \best{83.95$\pm$0.18} & 80.78$\pm$0.42& 77.23$\pm$0.72 & 80.33$\pm$0.29 & 80.65$\pm$0.59\\ 
        CIFAR10 & RESNET-56 & 77.98$\pm$0.24 & \best{80.98$\pm$0.48} & 79.66$\pm$0.26 &  77.32$\pm$0.35 & 77.97$\pm$0.45& 77.66$\pm$0.44\\
        \bottomrule
    \end{tabular}
    }
    
    \caption{Label smearing results under added label noise with probability of flipping each label $\rho=20\%$. Label smoothing parameter $\alpha=0.1$. For Patrini estimation of matrices for each label we use logits of an example falling into the $99.9$th percentile according to activations for that label. }
    \label{tbl:experiments_smearing}
\end{table*}

\begin{figure*}[!h]
    \centering
    \subfigure[Max logit, all data.]{%
    \includegraphics[width=0.48\textwidth]{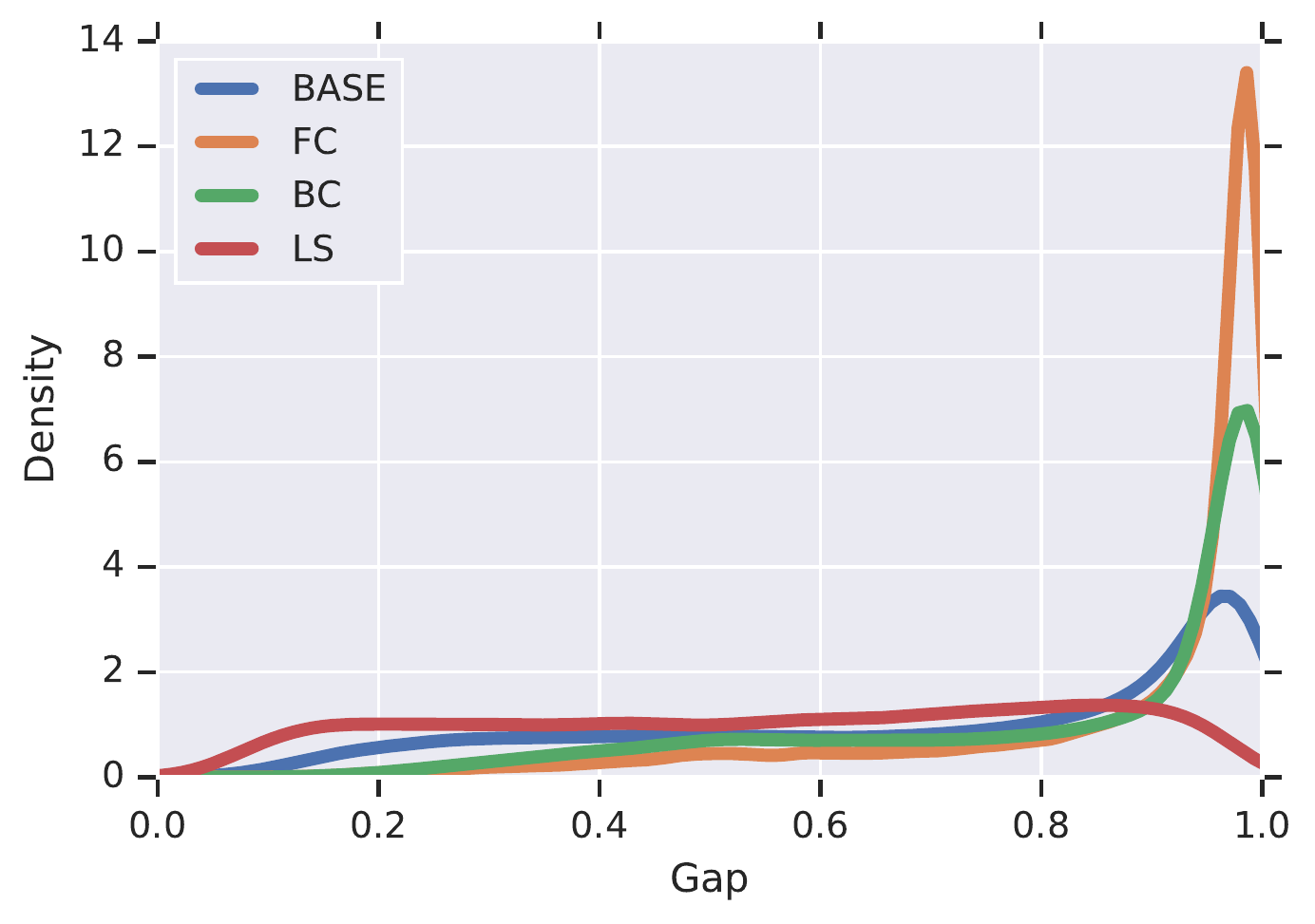}%
    }
    \subfigure[True label, non-noisy part.]{%
    \includegraphics[width=0.48\textwidth]{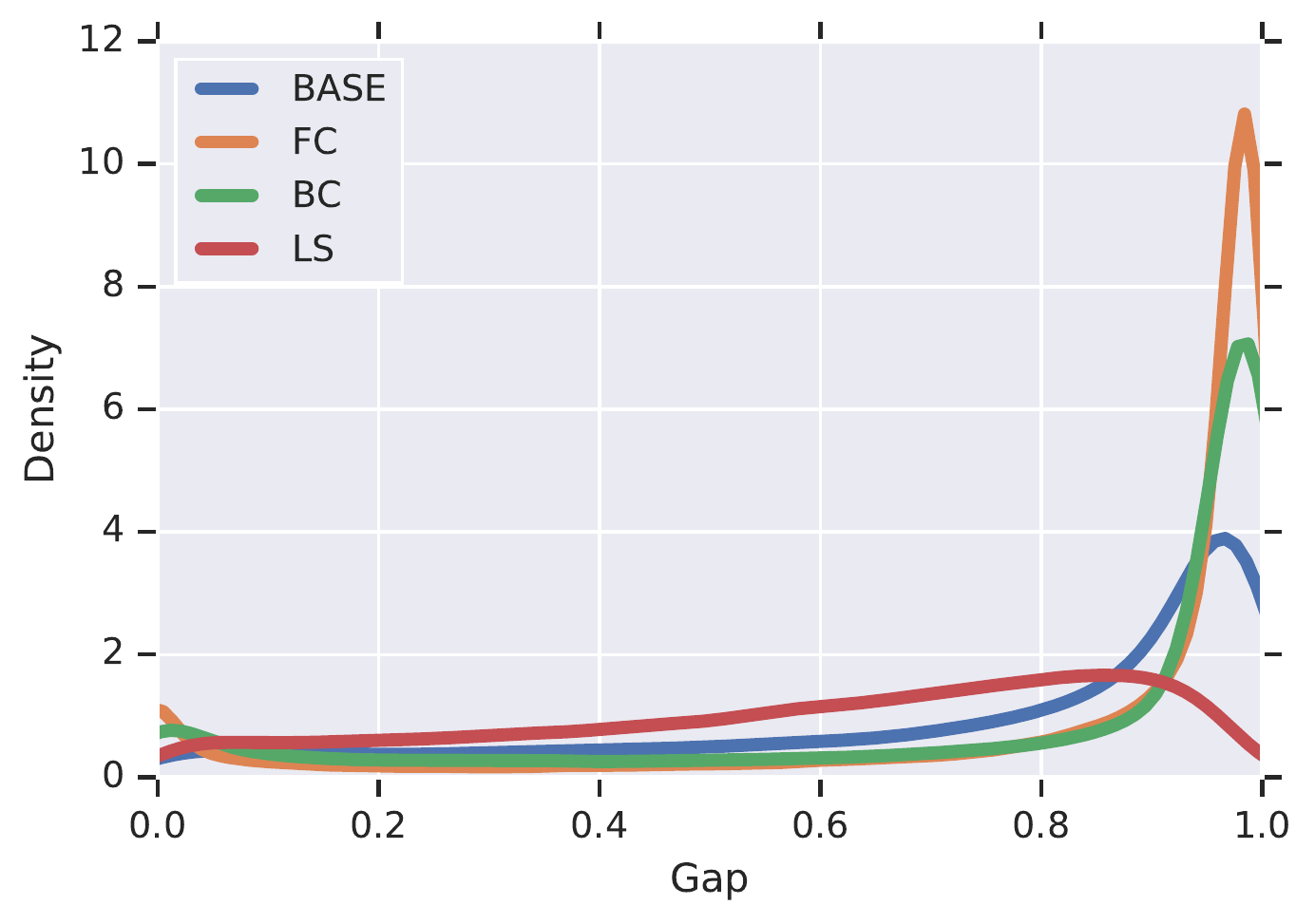}%
    }
    \caption{Density plots showing distribution differences between maximum logit (or corresponding to true label) and the average over all logits on different portions of train data and from different label smearing methods. Results using $\alpha=0.2$, dataset \cifarH{}, and the \resnetT{} model.}
    \label{fig:maxlogit_vs_avglogit}
\end{figure*}

Figure~\ref{fig:maxlogit_vs_avglogit} shows density plots of differences between maximum logit value (or corresponding to true/noisy label) and the average logit value across different portions of the training data.
We notice that while label smoothing is reducing the confidence (by lowering the peak around 1.0), backward correction and forward correction methods increase the confidence by boosting the spike. 
This is the case for both the noisy and true labels, however the effect is much stronger on the correct label logit activation.

\subsection{Logit visualisation plots}

\begin{figure*}[!t]
    \centering
    \subfigure[LS $\alpha=0$. Classes (0,1,2) \cifarH{}.]{%
    \includegraphics[width=0.24\textwidth]{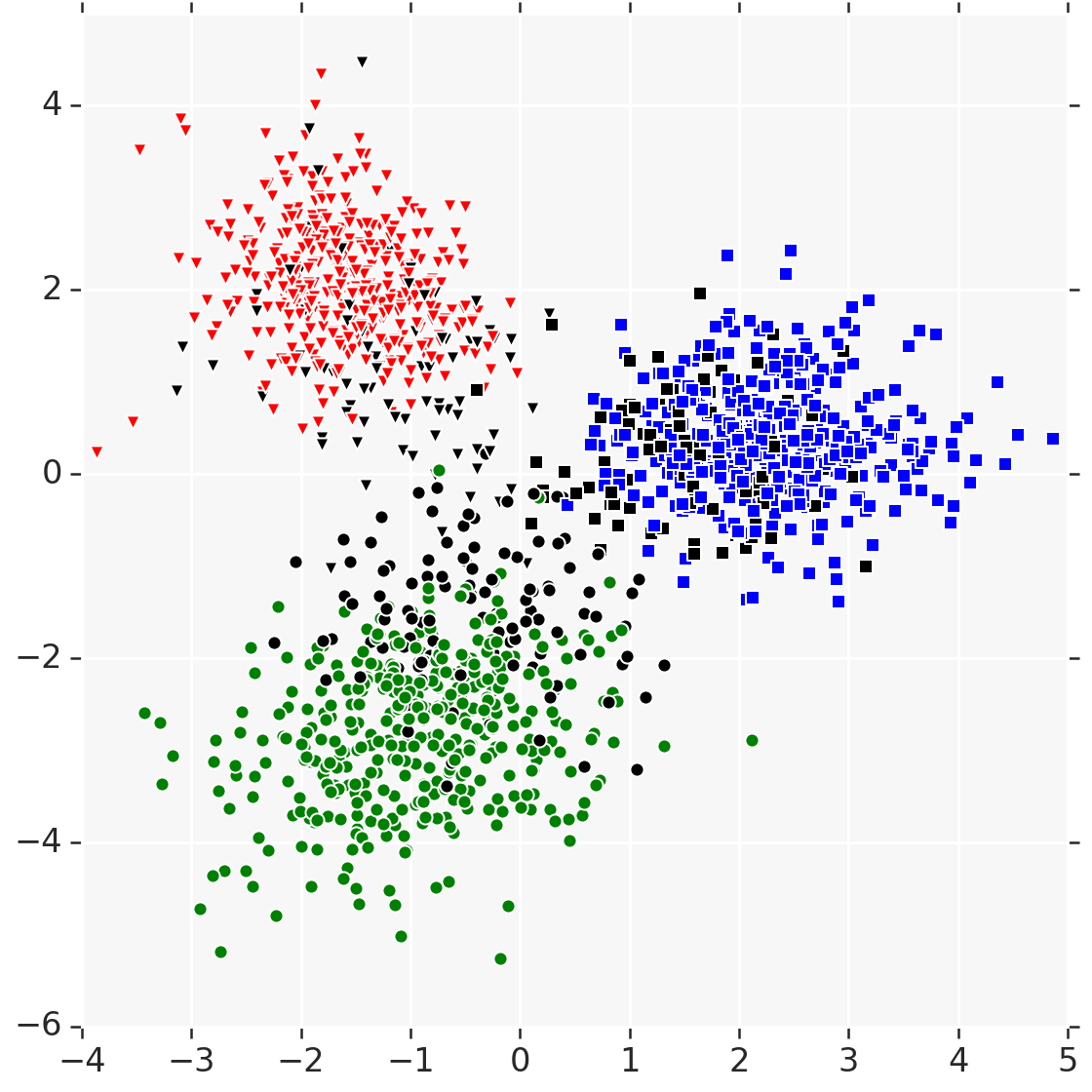}%
    }
    \subfigure[LS $\alpha=0.2$. Classes (0,1,2) \cifarH{}.]{%
    \includegraphics[width=0.24\textwidth]{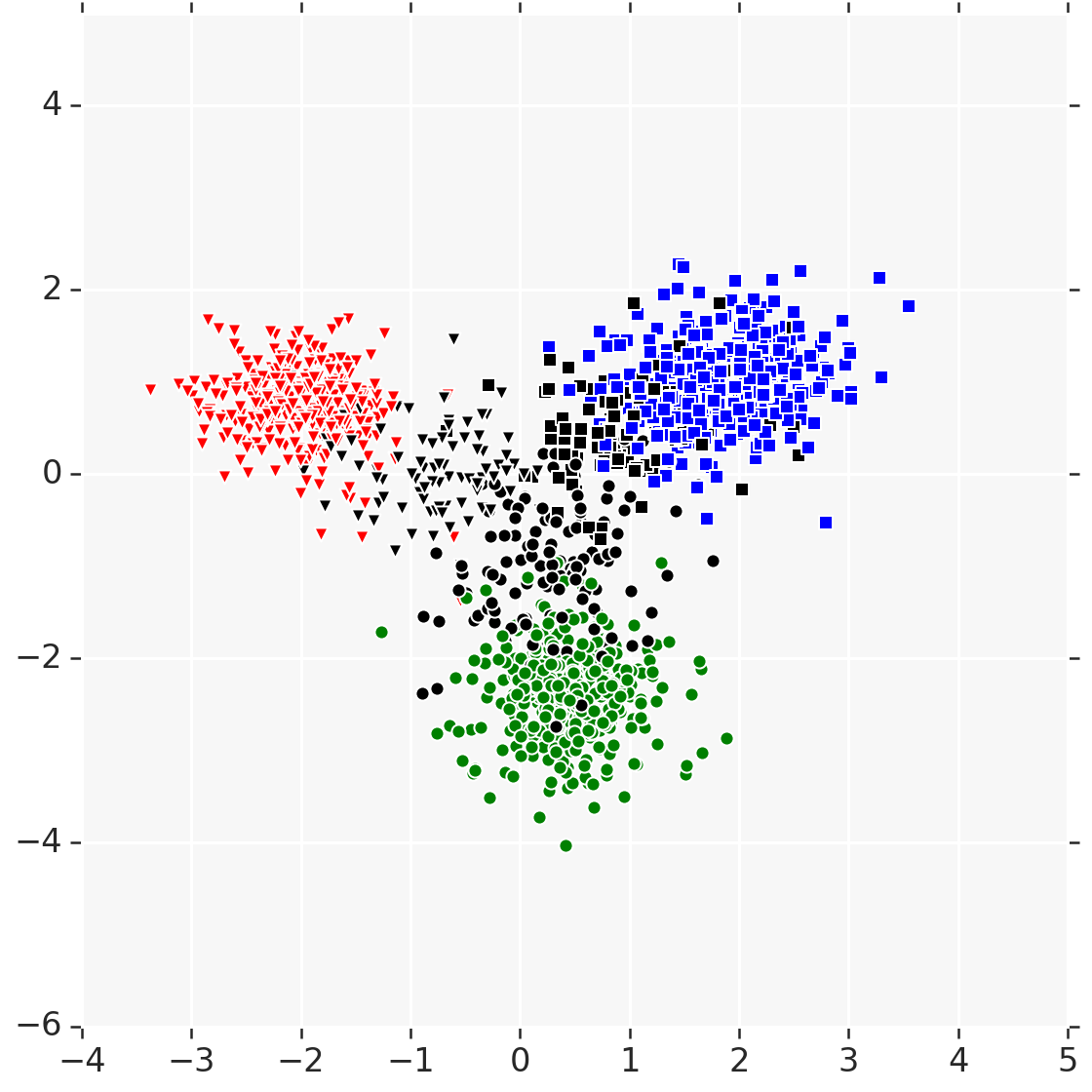}%
    }
    \subfigure[LS $\alpha=0$. Classes (3,4,5) \cifarH{}.]{%
    \includegraphics[width=0.24\textwidth]{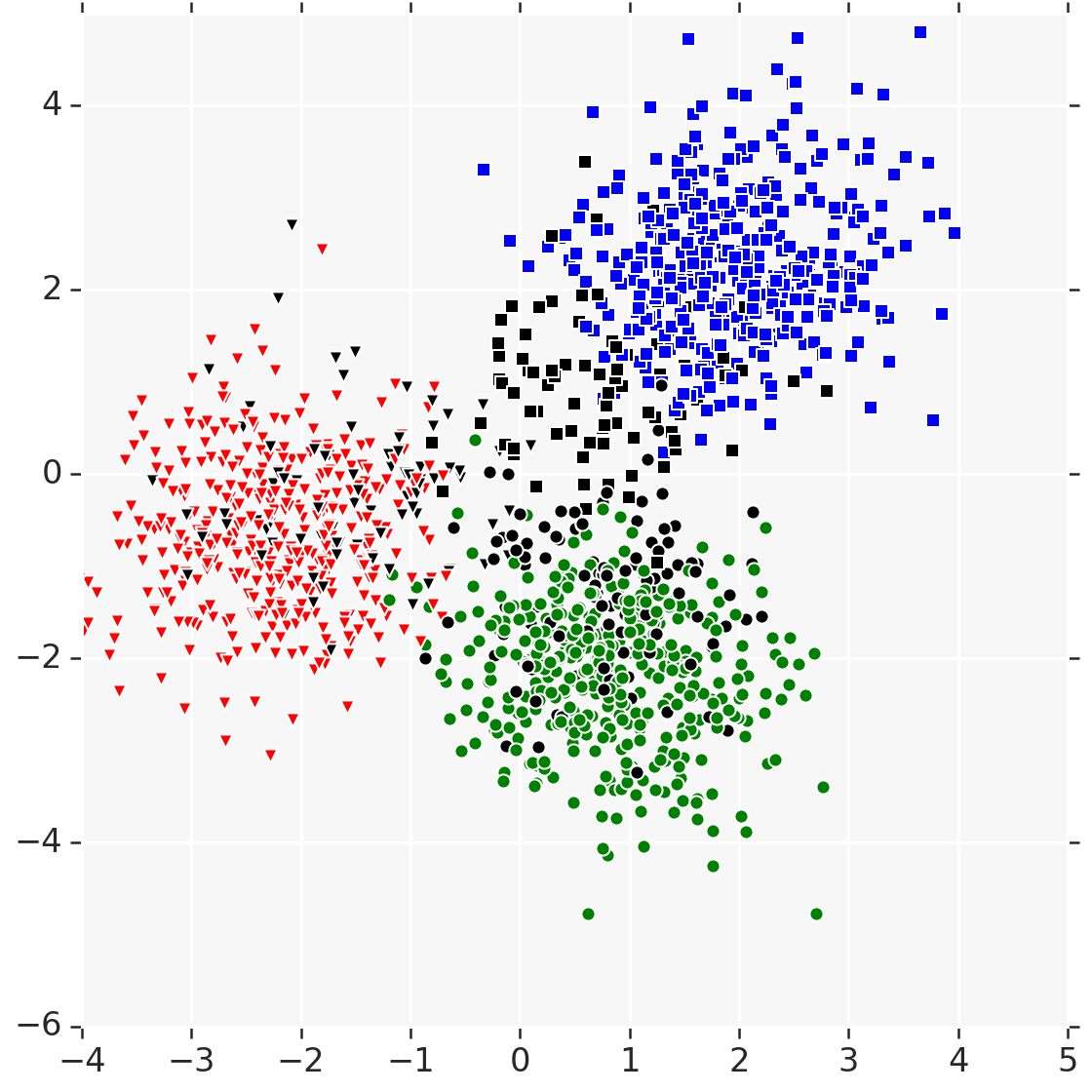}%
    }
    \subfigure[LS $\alpha=0.2$. Classes (3,4,5) \cifarH{}.]{%
    \includegraphics[width=0.24\textwidth]{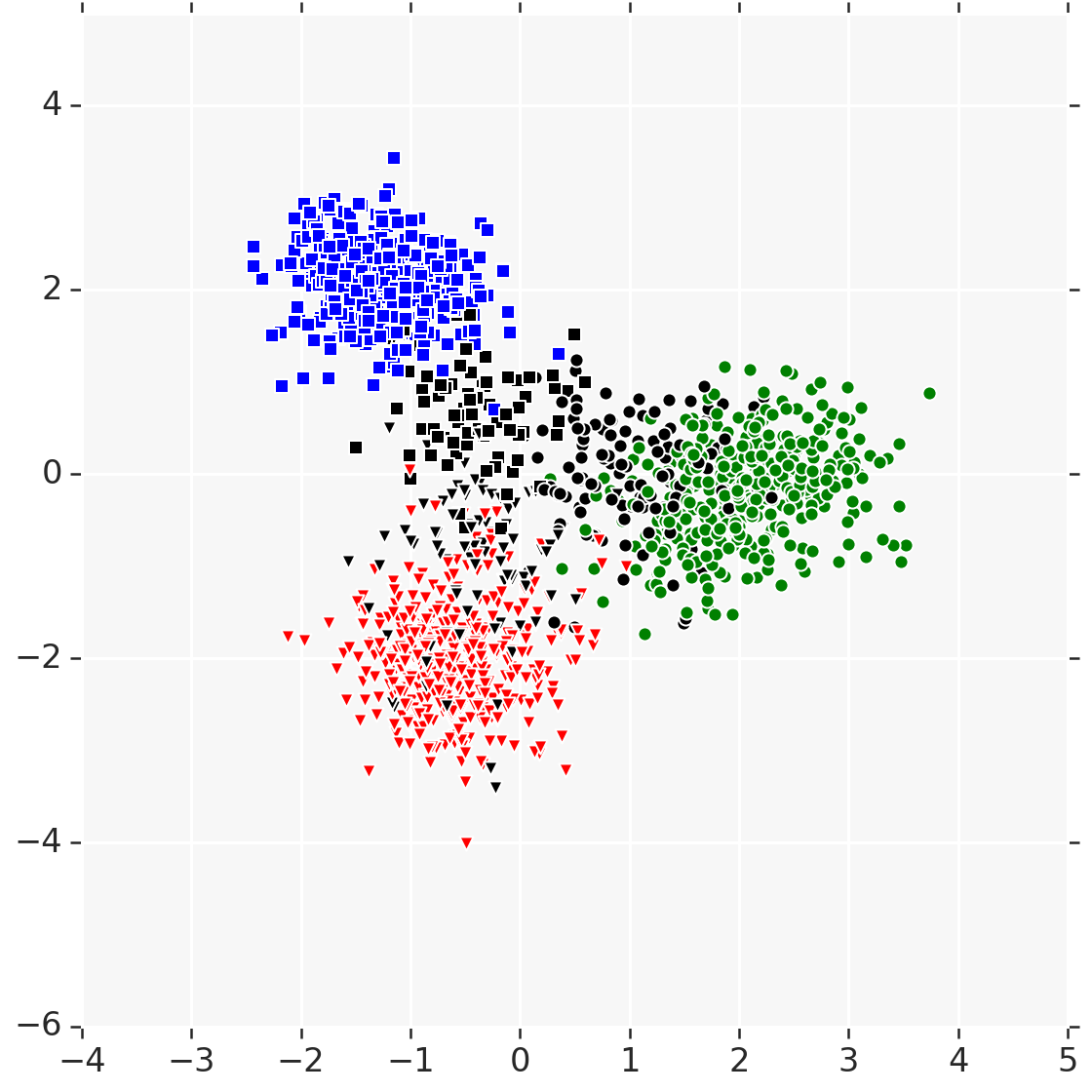}%
    }   

    \subfigure[LS $\alpha=0$. Classes (0,1,2) ~ \cifarT{}. ]{%
    \includegraphics[width=0.275\textwidth]{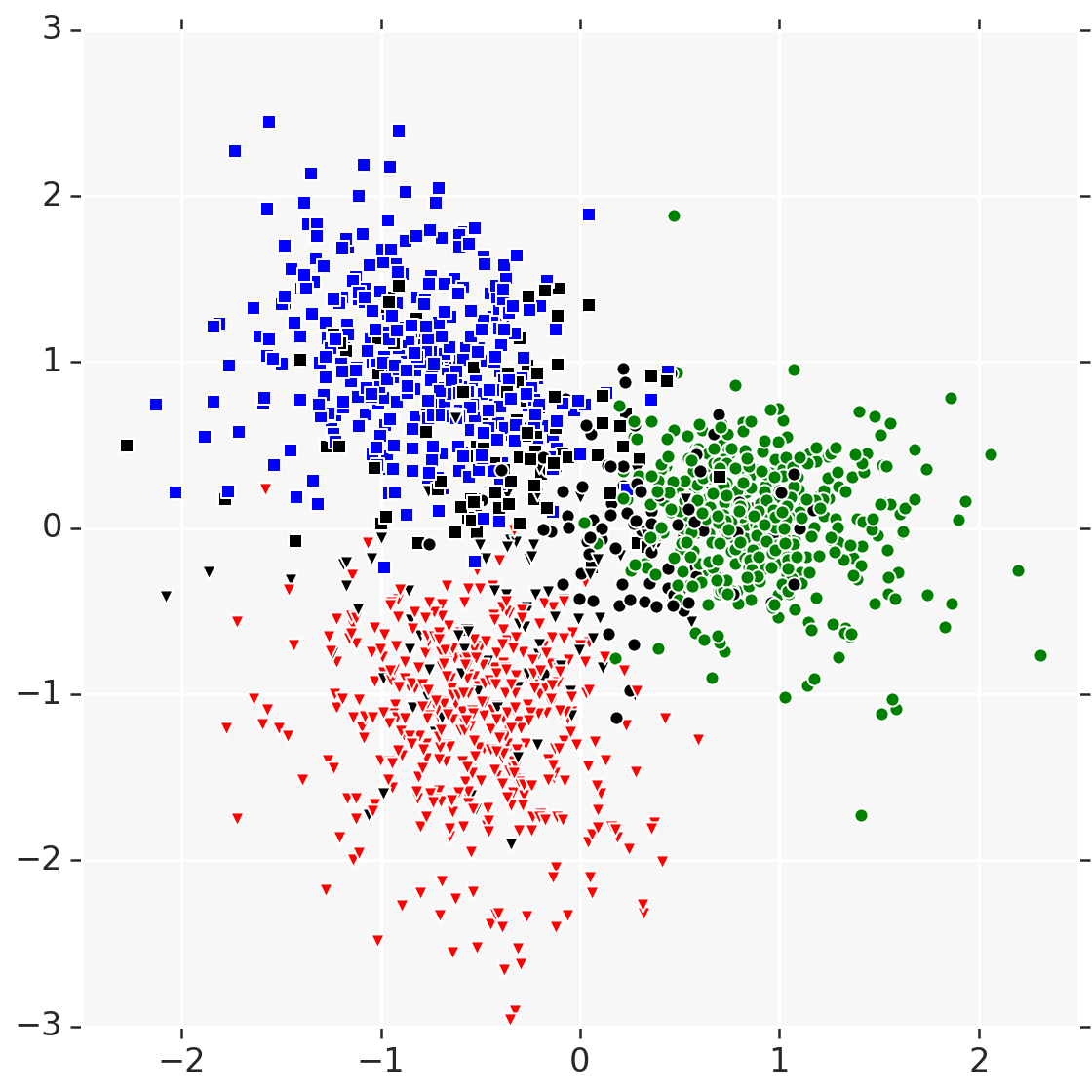}%
    }%
    \qquad
    \subfigure[LS $\alpha=0.2$. Classes (0,1,2) \cifarT{}.]{%
    \includegraphics[width=0.275\textwidth]{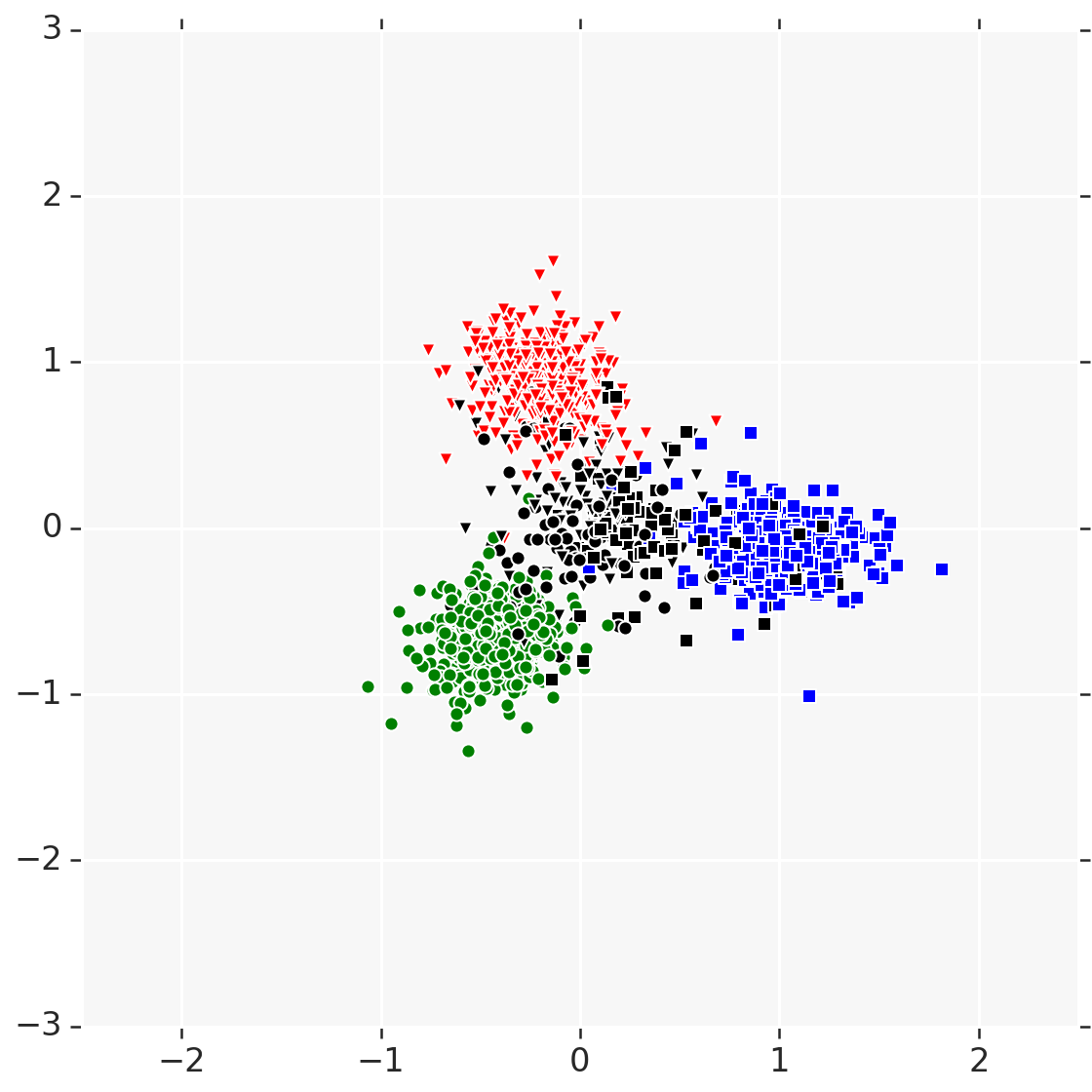}%
    }%
    \qquad
    \subfigure[LS $\alpha=0.7$. Classes (0,1,2) \cifarT{}.]{%
    \includegraphics[width=0.275\textwidth]{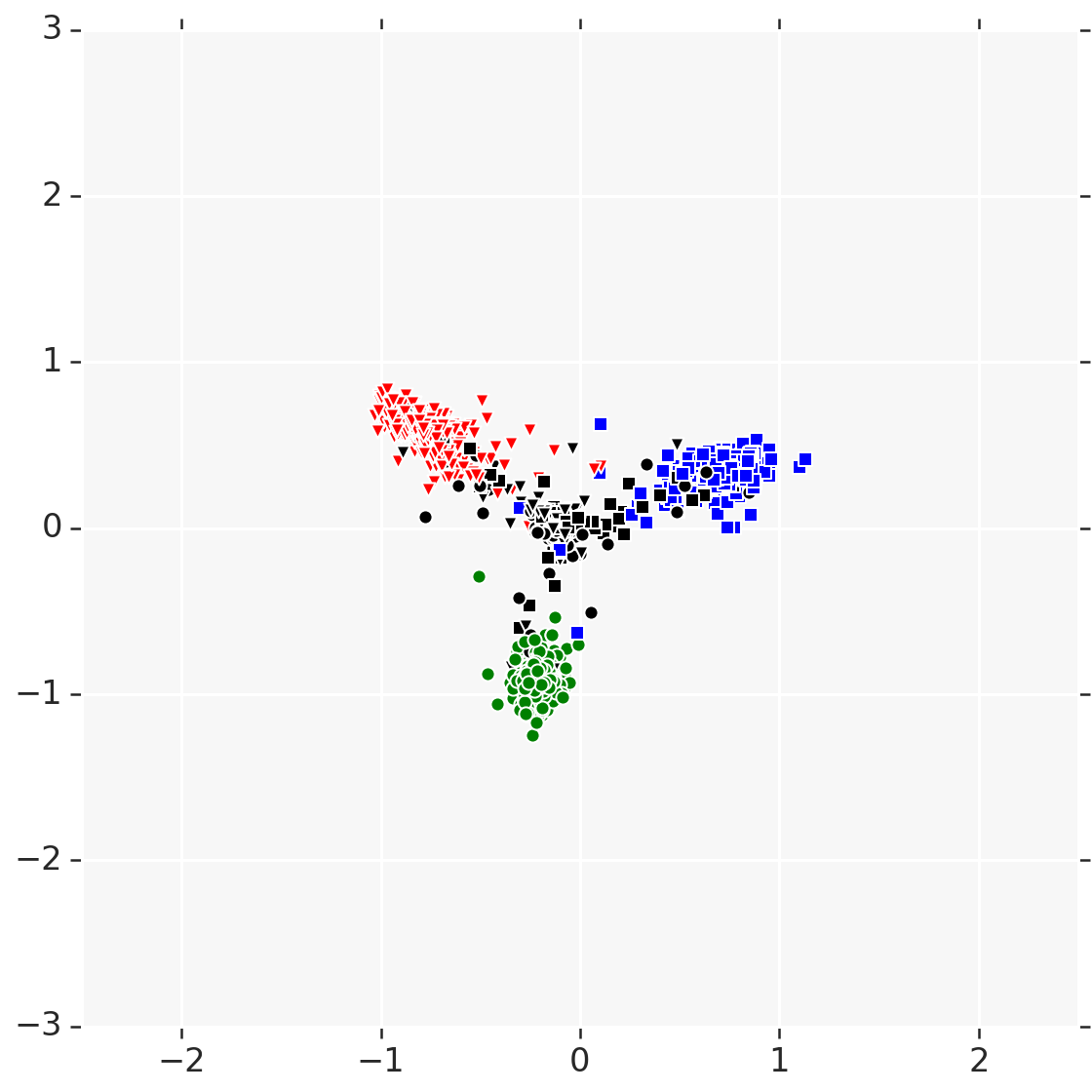}%
    }
    \caption{Effect of label smoothing on pre-logits (penultimate layer output) under label noise.
    Here, we visualise the pre-logits of a \resnetF{} for three classes on \cifarH{} (in the top figures), a \resnetT{} for three classes on \cifarT{} (in the bottom figures), using the procedure of~\citet{Muller:2019}.
    The black markers denote instances which have been labeled incorrectly due to noise.
    Smoothing is seen to have a denoising effect: the noisy instances' pre-logits become more uniform, and so the model learns to not be overly confident in their label.}
    \label{fig:smoothing_logits_r32_cif10}
\end{figure*}

\begin{figure*}[!t]
    \centering
    \subfigure[Forward correction $\alpha = 0.1$.]{%
    \includegraphics[width=0.24\textwidth]{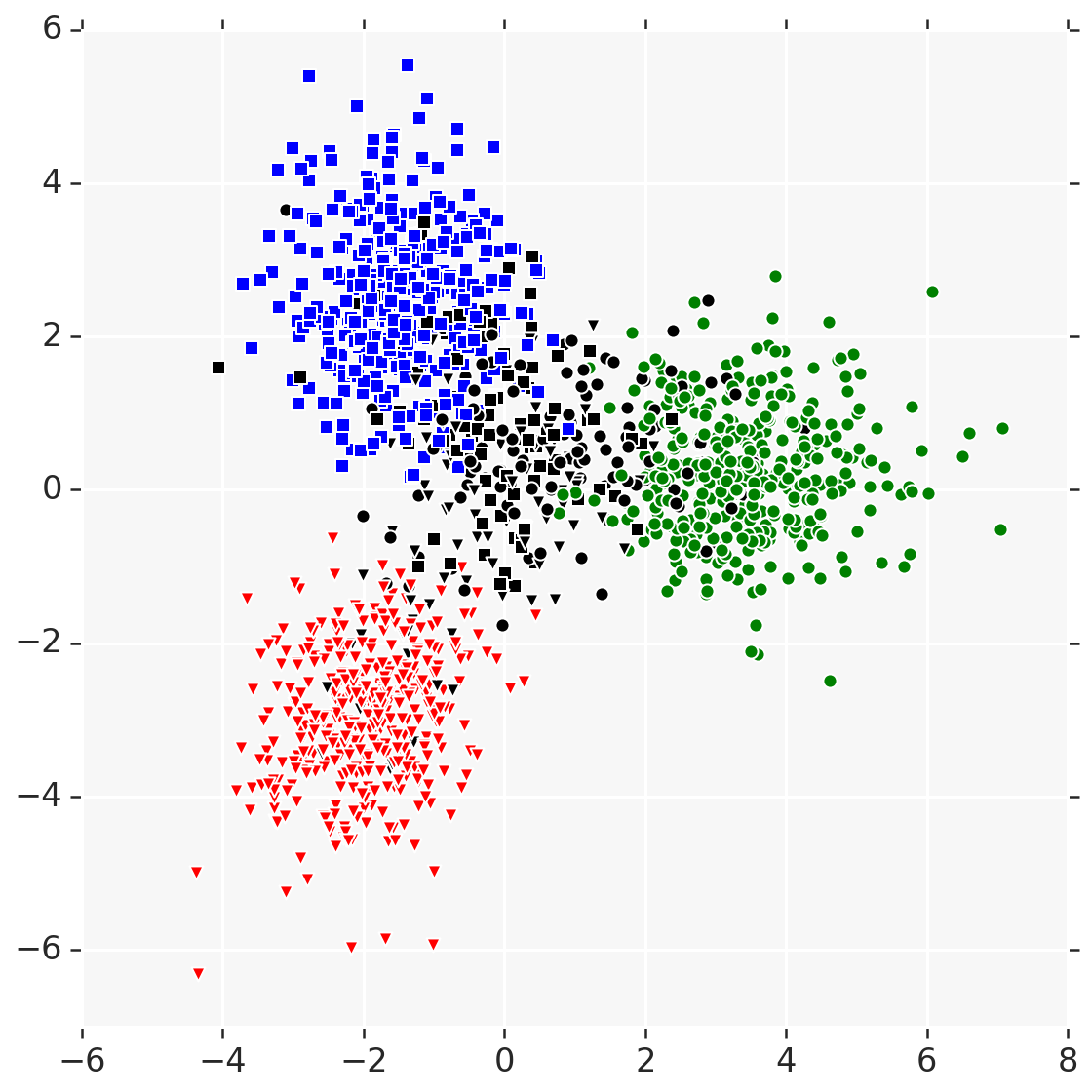}%
    }
    \subfigure[Forward correction $\alpha = 0.7$.]{%
    \includegraphics[width=0.24\textwidth]{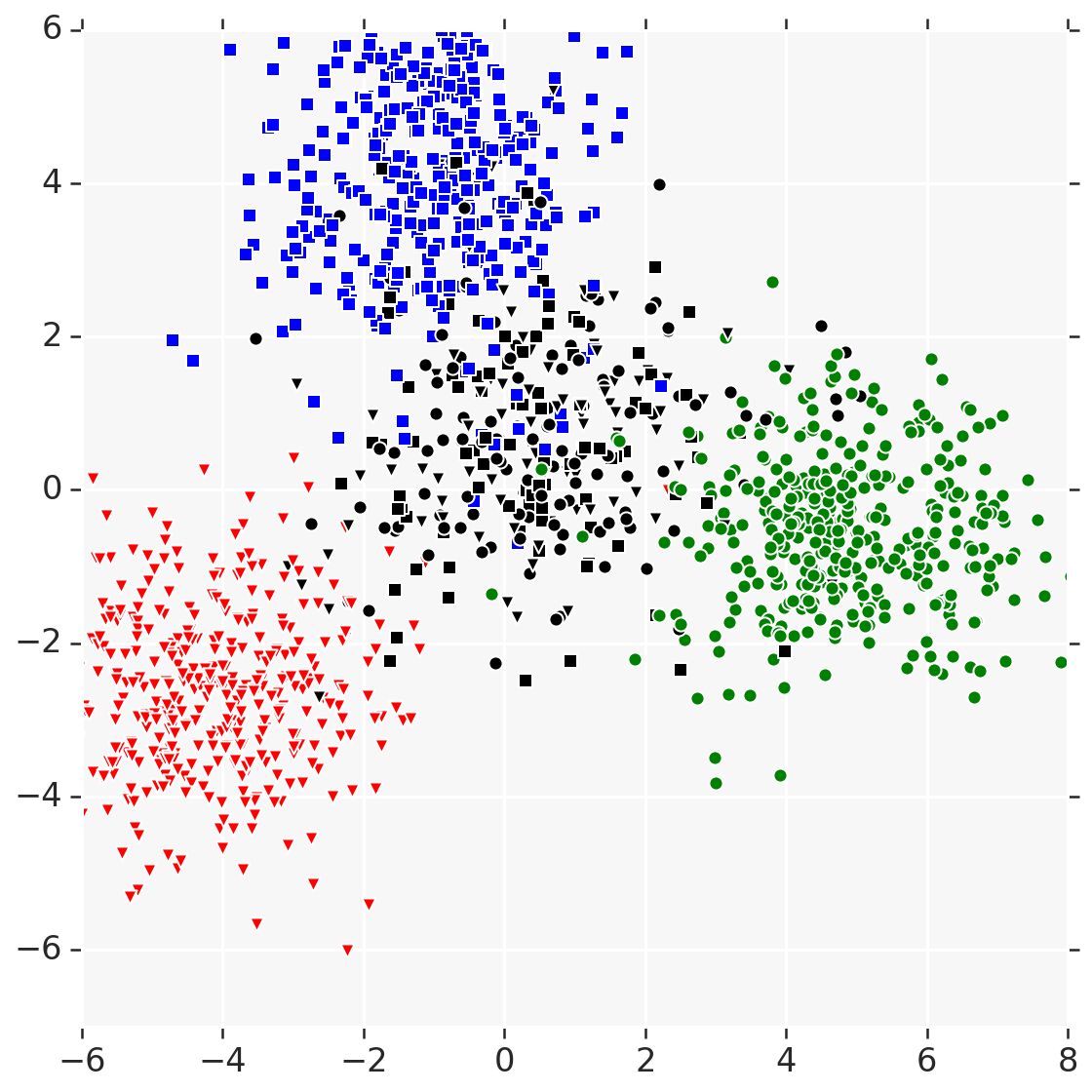}%
    }
    \subfigure[Backward correction $\alpha = 0.1$.]{%
    \includegraphics[width=0.24\textwidth]{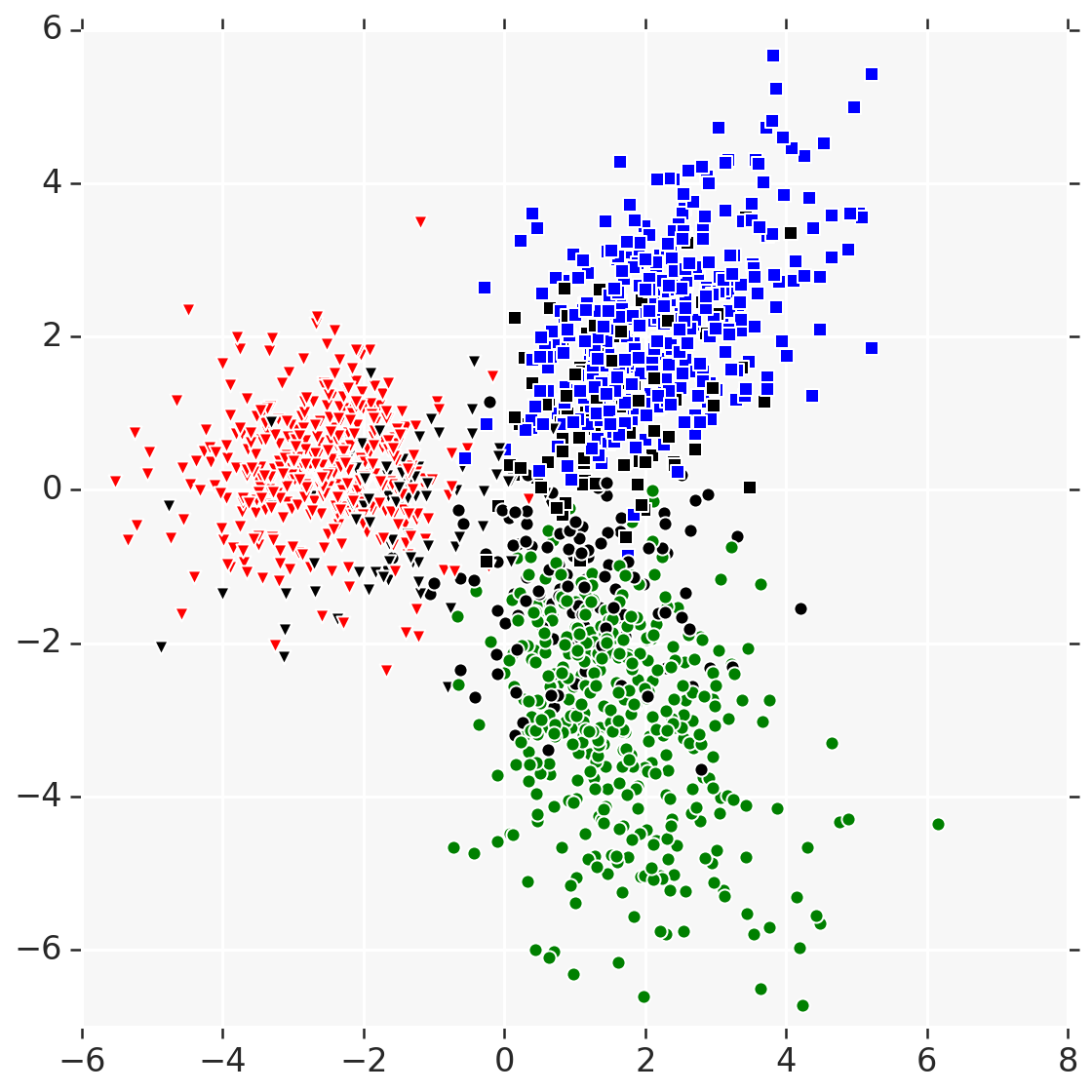}%
    }
    \subfigure[Backward correction $\alpha = 0.7$.]{%
    \includegraphics[width=0.24\textwidth]{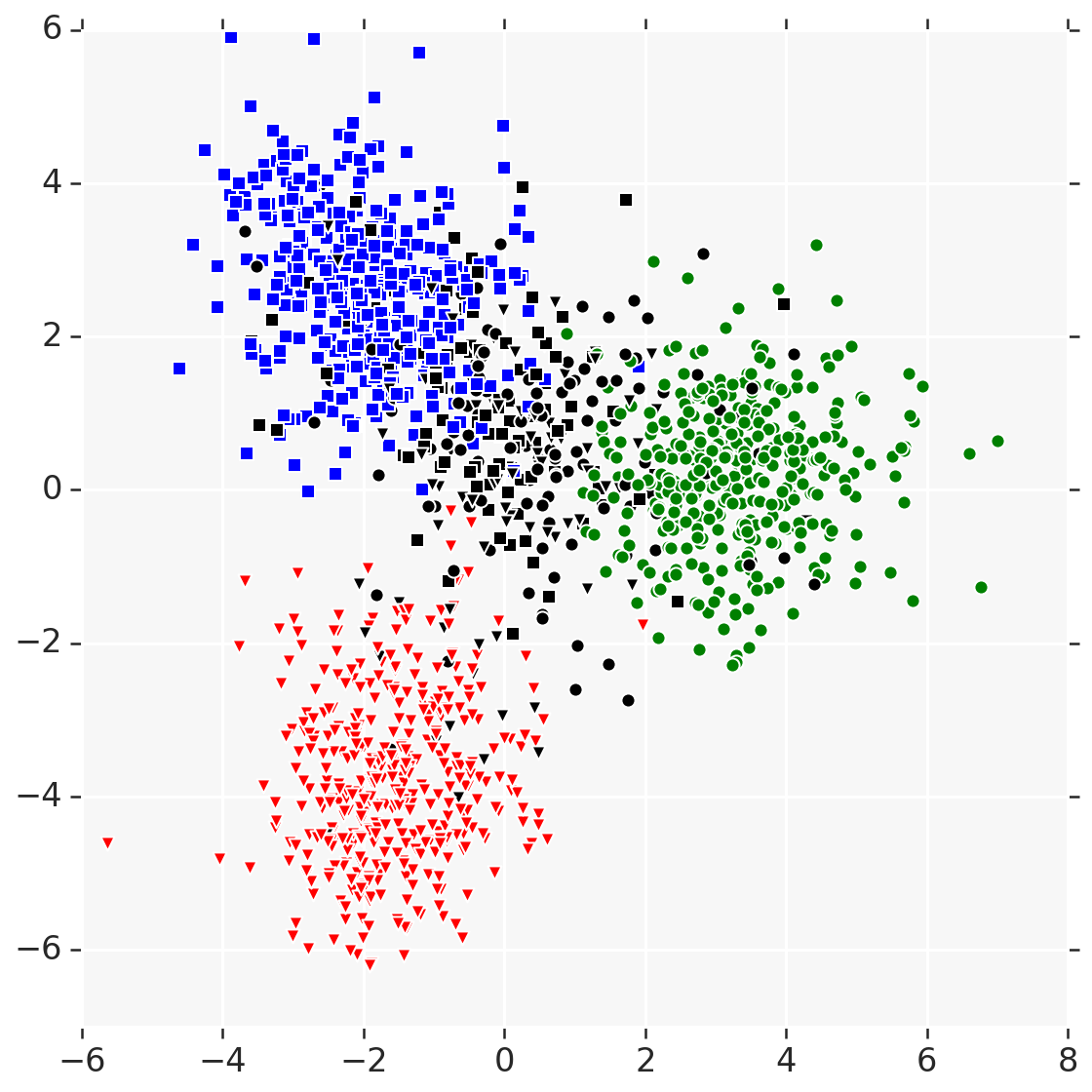}%
    }    
    \caption{Visualisation of logits for backward and forward correction of a \resnetT{} for three classes on \cifarH{}, using the procedure of~\citet{Muller:2019}. The red, blue and green colours denote instances from three different classes, and the black coloured points have label noise.}
    \label{fig:fb_logits}
\end{figure*}

In this section we present additional pre-logit visualization plots - for \cifarH{} trained with \resnetF{} in Figure \ref{fig:smoothing_logits_r32_cif10} (a-d), for \cifarT{} trained with \resnetT{} in Figure \ref{fig:smoothing_logits_r32_cif10}(e-g). Figure~\ref{fig:fb_logits} visualises the pre-logits for backward and forward correction on \cifarH{} trained with \resnetT{}. As before, we see that both methods are able to denoise the noisy instances.

